\documentclass[a4paper,fleqn]{cas-sc}

\usepackage[numbers]{natbib}

\usepackage{float}
\usepackage{algorithm} 
\usepackage{algpseudocode} 
\usepackage{amsmath}
\usepackage{subfigure}
\usepackage{multicol, latexsym, amsmath, amssymb}
\usepackage{blindtext}
\usepackage{subcaption}
\usepackage{caption}
\usepackage{soul}
\usepackage{pdflscape} 
\def\tsc#1{\csdef{#1}{\textsc{\lowercase{#1}}\xspace}}
\tsc{WGM}
\tsc{QE}

\begin{document}
\let\WriteBookmarks\relax
\def\floatpagepagefraction{1}
\def\textpagefraction{.001}

\shorttitle{cHM: A Universal Framework for Continuous Optimisation}    

\shortauthors{}  

\title [mode = title]{Constrained Hybrid Metaheuristic: A Universal Framework for Continuous Optimisation
}  

\author[add1,add2]{Piotr A. Kowalski}[orcid=0000-0003-4041-6900]
    \ead{pkowal@agh.edu.pl}\ead{pakowal@ibspan.waw.pl
    }

\author[add1,add3]{Szymon Kucharczyk} [orcid=0009-0002-2413-6984]
    \ead{kucharcz@agh.edu.pl} \cormark[1]
  
\author[add4, add5]{Jacek Ma{\'n}dziuk} [orcid=0000-0003-0947-028X]
    \ead{mandziuk@mini.pw.edu.pl}

\address[add1]{
Faculty of Physics and Applied Computer Science, 
AGH University of Krakow,\\
al. A. Mickiewicza 30, 30-059 Cracow, Poland.
}
\address[add2]{
Systems Research Institute, 
Polish Academy of Sciences,\\
ul. Newelska 6, 01-447 Warsaw, Poland.
}
\address[add3]{
AGH Doctoral School, 
AGH University of Krakow,\\
al. A. Mickiewicza 30, 30-059 Cracow, Poland.
}
\address[add4]{
Faculty of Mathematics and Information Science, 
Warsaw University of Technology \\
Koszykowa 75, 00-662 Warsaw, Poland.
}
\address[add5]{
Faculty of Computer Science, 
AGH University of Krakow,\\
al. A. Mickiewicza 30, 30-059 Cracow, Poland.
}

\cortext[1]{Corresponding author}




\begin{abstract}
This paper presents the constrained Hybrid Metaheuristic (cHM) algorithm as a general framework for continuous optimisation. Unlike many existing metaheuristics that are tailored to specific function classes or problem domains, cHM is designed to operate across a broad spectrum of objective functions, including those with unknown, heterogeneous, or complex properties such as non-convexity, non-separability, and varying smoothness.
We provide a formal description of the algorithm, highlighting its modular structure and two-phase operation, which facilitates dynamic adaptation to the problem's characteristics. A key feature of cHM is its ability to harness synergy between both candidate solutions and component metaheuristic strategies. This property allows the algorithm to apply the most appropriate search behaviour at each stage of the optimisation process, thereby improving convergence and robustness.
Our extensive experimental evaluation on 28 benchmark functions demonstrates that cHM consistently matches or outperforms traditional metaheuristics in terms of solution quality and convergence speed. In addition, a practical application of the algorithm is demonstrated for a feature selection problem in the context of data classification. The results underscore its potential as a versatile and effective black-box optimiser suitable for both theoretical research and practical applications.
\end{abstract}




\maketitle

\section{Introduction}
\label{sec:introduction}
Optimisation is a foundational element across a wide array of scientific and engineering disciplines. It serves as the mathematical framework behind a myriad of practical tasks: designing efficient transportation networks \cite{MANDZIUK2017INFSCI}, fine-tuning hyperparameters in machine learning models \cite{schaipp2025surprising,Sieradzki2025EATPE}, scheduling resources in industrial processes \cite{WALEDZIK2018INFSCI}, and solving complex problems in physics and biology \cite{butenko2014numerical}. 

Many modern optimisation problems are characterised by high dimensionality, nonlinearity, and non-convexity, leading to rugged landscapes populated with numerous local minima. Traditional optimisation techniques, particularly those based on convex assumptions, often fail in such settings. Furthermore, the need for algorithms that can handle noisy, incomplete, or non-differentiable objective functions is increasingly common in applications such as deep learning, financial modelling, and network design \cite{bubeck2015convex,kowalski2016training}.

Recent developments in metaheuristic optimisation have increasingly focused on enhancing algorithmic adaptability, scalability, and real-world applicability. One prominent direction is the design of adaptive and self-configuring algorithms, which dynamically adjust control parameters during the optimisation process to better respond to problem characteristics \cite{Wagner2009,Okulewiczetal2020,Okulewiczetal2022}. These methods aim to reduce manual tuning and improve algorithm robustness across heterogeneous landscapes.

Another significant trend is the integration of learning mechanisms into metaheuristics, including surrogate modelling and reinforcement learning, to guide search more effectively, especially in expensive black-box scenarios \cite{jin2011surrogate, wang2021survey}. These hybrid approaches often strike a balance between global exploration and cost-aware local exploitation.

In parallel, hyper-heuristics and algorithm portfolios have emerged as high-level strategies that aim to automatically select or generate appropriate heuristics during the optimisation process, based on problem features or past performance \cite{Burke2013,Zakrzewski2025KES}. Such strategies reflect a shift from designing individual algorithms to constructing flexible frameworks that can adapt to a wide range of tasks.

The applications of metaheuristics have also expanded, ranging from engineering optimisation \cite{yang2010engineering} and energy systems \cite{GUSAIN2023101711} to bioinformatics and logistics \cite{OKULEWICZ2019SWEVO,kumar2021natureinspiredoptimizationalgorithmsresearch}. These developments underscore the growing importance of general-purpose, flexible optimisation methods—an area where the proposed cHM procedure is particularly well-positioned.

Overcoming these challenges requires innovative approaches that go beyond the limitations of classical methods. Although gradient-based methods are efficient for smooth and convex problems, they are prone to getting caught in local minima and often rely on stringent smoothness conditions \cite{ruder2016overview}. Metaheuristic algorithms like Simulated Annealing (SA) \cite{siddique2016simulated} and Particle Swarm Optimisation (PSO) \cite{wang2018particle} have been introduced to address some of these issues by incorporating stochastic or population-based strategies. However, these methods can suffer from slow convergence and require careful tuning of hyperparameters.


A key principle underlying many nature-inspired metaheuristic optimisation algorithms is the concept of synergy among individuals. In these algorithms, a population of candidate solutions (often referred to as individuals, particles, or agents) collectively explores the search space. Synergy arises when the interactions among individuals-whether through information sharing, imitation, or competition-lead to emergent behaviours that surpass the capabilities of isolated search processes.

In PSO, for example, individuals adjust their trajectories based on both their own experience and the experiences of their neighbours, enabling faster convergence to optimal regions of the search space \cite{pso_origin,lukasik2014fully}. Similarly, in Evolutionary Algorithms (EAs), crossover and mutation operators combine genetic material from different individuals to generate offspring with potentially superior traits \cite{eiben2015introduction}

In this landscape, hybrid metaheuristic approaches, which combine the strengths of multiple optimisation strategies, have gained increasing attention. By leveraging the complementary properties of diverse metaheuristics, hybrid algorithms can enhance both exploration and exploitation capabilities, leading to more efficient optimisation. This motivates the exploration of the constrained Hybrid Metaheuristic (cHM) algorithm as a universal framework for continuous optimisation, capable of handling various complex, real-world objective functions beyond specific applications.

\subsection{Motivation}
A key principle underlying many nature-inspired metaheuristic optimisation algorithms is the synergy of individuals. In these algorithms, a population of candidate solutions cooperates by sharing information, imitating successful strategies, or competing for better positions in the search space. This interaction creates an emergent behaviour where the group collectively explores and exploits the search space more efficiently than isolated individuals could. Synergy improves the balance between diversification and intensification of the search, leading to faster convergence and a greater likelihood of finding high-quality solutions in complex and multimodal landscapes.

The proposed cHM procedure not only leverages the fundamental idea of synergy among individuals but also introduces a higher-level synergy by combining multiple metaheuristics within a single framework. By allowing different optimisation strategies to cooperate and adapt dynamically, cHM fosters an emergent collaboration between algorithms themselves, not just their constituent individuals. This multi-layered synergy enhances the algorithm's versatility and robustness, enabling it to tackle a broader range of complex optimisation problems. Such an approach opens new possibilities for creating more adaptive and efficient optimisation processes, offering significant advantages over traditional single-method metaheuristics.

Moreover, the proposed cHM does not introduce yet another algorithm based on specific natural phenomena or behaviours. Instead, it offers a universal optimisation procedure that flexibly adapts to a wide range of functions and problem landscapes. By not tying itself to the mechanisms of a single nature-inspired model, cHM avoids the limitations of specialised algorithms and provides a general-purpose framework suitable for continuous, high-dimensional optimisation tasks. This universality makes it an attractive choice for applications where problem structures are diverse and not known in advance.

In many real-world optimisation problems, the objective function is not known explicitly and exhibits highly heterogeneous properties across different regions of the search space. Some areas may be smooth and convex, while others may be rugged or even discontinuous. As no single algorithm performs optimally under all such conditions, the proposed method introduces a cooperative framework in which diverse metaheuristics interact. This allows the system to dynamically exploit the strengths of each component method during different phases of the optimisation process, resulting in a self-adaptive, higher-level synergy between algorithms. 

\subsection{Contribution}

The primary contributions of this paper are as follows.

\begin{itemize}
\item We present a novel \textit{constrained Hybrid Metaheuristic} (cHM) algorithm---a universal optimisation framework for continuous, high-dimensional problems with unknown or heterogeneous objective functions. 

\item We introduce and formalise the concept of synergy not only among individuals within a population, but also across different metaheuristic algorithms operating within a unified architecture.

\item We demonstrate that cHM dynamically leverages the strengths of multiple optimisation strategies, assigning them adaptively to different phases of the optimisation process based on empirical performance.

\item We conduct a comprehensive experimental study on 28 standard benchmark functions, showcasing the generalisation ability and competitiveness of the method compared to single-strategy metaheuristics.

\item We position cHM as a flexible foundation for future research in hybrid, adaptive, and self-configuring optimisation systems.

\item We show a practical application of the cHM algorithm in a feature selection problem in the context of tabular data classification.
\end{itemize}

\subsection{Structure of the Paper}
The remainder of this paper is organised as follows. Section~\ref{sec:related_work} reviews related work on hybrid metaheuristics, summarising prior efforts to combine multiple optimisation strategies for improved performance. Section~\ref{sec:cHM} presents the proposed cHM algorithm, detailing its core ideas, design principles, and algorithmic description. Section~\ref{sec:experimental_setup} outlines the experimental setup, including benchmark functions, parameter settings, and performance evaluation criteria. Section~\ref{sec:results} discusses the experimental results, comparing cHM to traditional metaheuristic algorithms across various metrics. Section~\ref{sec:Application} presents an example of cHM application to feature selection in machine learning classification. Finally, Section~\ref{sec:Conclusions} concludes the paper and highlights potential directions for future research.

\section{Related work}
\label{sec:related_work}

Research in metaheuristic optimisation has developed along several complementary directions that aim to improve robustness, scalability, and applicability across heterogeneous problem landscapes. A first important line of work is based on the notion of algorithm portfolios and hyper-heuristics. The algorithm selection paradigm introduced by Rice \cite{Rice1976} and later reinforced by the No Free Lunch theorems \cite{WolpertMacready2002} highlights that no single algorithm can dominate across all problem classes. This observation motivated the development of portfolio approaches such as SATzilla, which demonstrated the effectiveness of selecting algorithms per instance using empirical hardness models \cite{SATzilla2008}. Recently, portfolio-based approaches relying on surrogate management methods have been proposed by \cite{Zakrzewski2025KES}. Beyond portfolios, hyper-heuristics provide higher-level search methodologies that operate on a space of heuristics rather than candidate solutions themselves. Comprehensive surveys document both selection-based and generation-based hyper-heuristics and their applicability to diverse optimisation domains \cite{Burke2013, Dokeroglu2024}. Metaheuristic optimisation methods have also been applied in the domain of large language models (LLMs), particularly in contexts involving automated hyperparameter tuning or the generation of heuristics guiding the learning process \cite{Huang2023LLMOpt}. For instance, \cite{vanStein2024} proposed a hybrid approach that combines LLMs with HPO procedures, separating the generation of algorithmic structures from parameter tuning. Another example is AutoEP: LLM-driven Automation of Hyperparameter Design, where chains of reasoning produced by multiple LLMs interpret information about the search landscape and adapt the hyperparameters of metaheuristics in an online fashion \cite{AutoEP2025}. These works illustrate a steady move from hand-crafted algorithms towards flexible frameworks that adaptively orchestrate existing methods.

Another significant trend has been the hybridisation of metaheuristics. Early taxonomies described how different strategies could be combined to exploit complementary strengths in balancing exploration and exploitation \cite{BlumRoli2011, TalbiTaxonomy}. Hybridisation also underpins island models and cooperative co-evolution, where populations or subcomponents evolve in parallel while exchanging information \cite{Luke2011CCEA, Dorronsoro2013}.

Island-based algorithms are widely recognised as an effective framework for structuring population-based search. Early studies demonstrated that island models can mitigate premature convergence and improve exploration by maintaining separated subpopulations \cite{Whitley1999}. Subsequent analyses examined the impact of migration size and frequency, highlighting the delicate balance between exploration and exploitation required to achieve robust performance \cite{Skolicki2005a, Skolicki2005b}. Later work extended these insights through dynamic island models based on spectral clustering \cite{Meng2017}. Variants of the island framework have also been studied in the context of different heuristics, including PSO \cite{Abadlia2017}, Differential Evolution (DE) \cite{Apolloni2008}, and Ant Colony Optimisation (ACO) \cite{Mora2013}, confirming the generality and flexibility of the island model as a platform for parallel metaheuristics. Recent approaches emphasise adaptive migration and decomposition, showing advantages in large-scale and multi-objective optimisation. A diversity-driven cooperating portfolio proposed by \cite{Zychowski2025KES,ZychowskiGECCO2025,Zychowski2025Zakopane} explicitly analyses how diversity can guide the exchange of information across heterogeneous metaheuristics. This line of research highlights the importance of synergy not only within populations but also between distinct algorithms. In a similar spirit, \cite{Zajecka20245775} applies a portfolio of optimisation metaheuristics in an island-based setup to solve transportation problems.

Hybridisation taxonomies further clarify how cooperative mechanisms can be incorporated into metaheuristics. Parallel hybrid island approaches illustrate how heterogeneous methods can be embedded within a unified structure \cite{Li2022}. Other studies have investigated design choices in cluster geometry optimisation, providing empirical evidence that topology and migration influence both effectiveness and diversity in island models \cite{Leitao2015}. Further frameworks introduced co-evolutionary and adversarial portfolio construction, showing that algorithmic diversity can be harnessed through systematic interaction \cite{Tang2021, Liu2020}. The above frameworks position island-based cooperation not only as a technique for parallelisation but also as a means of systematically combining complementary strengths across heterogeneous algorithms. 

Diversity preservation has emerged as a critical theme in distributed evolutionary computation. Alternative island models have been proposed to promote diversity by maintaining niches across subpopulations \cite{Gustafson2006}, while hierarchical frameworks aimed to sustain long-term search potential \cite{Hu2005}. Dual migration mechanisms explicitly preserve diversity through interactions between both neighbouring and distant islands \cite{Gozali2019}, and migration strategies guided by diversity metrics have been shown to reduce stagnation \cite{Araujo2023}. A survey of distributed evolutionary algorithms emphasises that diversity is a foundation of robust performance \cite{Gong2015}. Taken together, these works demonstrate that the effectiveness of hybrid portfolios depends not only on algorithmic cooperation but also on carefully managing the diversity that enables such cooperation to flourish.

In parallel, parameter control and automated configuration have been studied as mechanisms for enhancing the adaptability of optimisation methods. A systematic review in. \cite{Karafotias2014} classifies deterministic, adaptive and self-adaptive schemes for parameter control, illustrating how online adaptation can substantially improve robustness. Complementary to this, automated configuration techniques such as SMAC \cite{Hutter2011SMAC} optimise parameters offline across distributions of problem instances. A survey of algorithm configuration methods \cite{Kotthoff2016} and subsequent frameworks like AutoFolio demonstrate the potential of combining configuration with portfolio selection, thereby reinforcing the trend towards automation in algorithm design.

Another related research line is the incorporation of learning mechanisms into metaheuristics. Surrogate-assisted evolutionary computation has received considerable attention as a strategy for dealing with expensive black-box functions. \cite{jin2011surrogate} surveys early developments in this area, while more recent work confirms that surrogate modelling can dramatically reduce computational cost while maintaining global search capability \cite{liu2024large}; \cite{cho2017survey}. Reinforcement learning has also been integrated as a meta-level controller to adapt operator choice or heuristic selection in dynamic search environments \cite{Seyyedabbasi2023, BolufeRohler2025}. Together, these learning-enhanced methods illustrate how machine learning can strengthen the exploratory and exploitative balance of metaheuristics, particularly in heterogeneous landscapes.

The constrained Hybrid Metaheuristic algorithm aligns with several of these developments. It shares conceptual ground with portfolio and hyper-heuristic approaches by orchestrating multiple optimisers within a single framework. At the same time, it inherits ideas from hybrid and island models by fostering synergy between heterogeneous algorithms. Unlike many portfolio methods that select a single solver for each instance, cHM emphasises dynamic cooperation during a single optimisation run, allocating different strategies to phases of the search process according to their empirical performance. This phase-wise collaboration, combined with explicit constraint handling, positions cHM as a general-purpose and algorithm-agnostic framework for continuous optimisation, rather than as yet another algorithm inspired by a particular natural process.

It is important to distinguish this work from the first published application of cHM \cite{CHM}, which was focused on probabilistic neural networks. That earlier study demonstrated the utility of cHM for selecting and refining smoothing parameters in a specific neural model and validated the approach across a collection of benchmark datasets. In contrast, the present article moves beyond a task-specific demonstration and undertakes a systematic investigation of cHM as a general optimiser. In particular, we benchmark it on a broad suite of continuous test functions, formalise its multi-phase cooperative structure and constraint handling mechanisms, and analyse the contribution of synergy across algorithms through controlled ablations. While the PNN-focused study positioned cHM as a tool for machine learning, the current paper situates it within the broader landscape of metaheuristic research, highlighting its potential as a universal and reusable optimisation framework applicable to diverse and heterogeneous problem settings.

\section{cHM Algorithm}
\label{sec:cHM}

Computational optimisation is a rapidly developing field of research in computer science. One class of problem solvers are metaheuristics methods, especially swarm-based techniques. Despite their general strength in nonparametric optimisation and high adaptability to various problems, their performance may be satisfactory only in certain areas. 

Here, we propose a \textit{constrained Hybrid Metaheuristic} algorithm that leverages the advantages of several swarm-based optimisation techniques in a hybrid method. The cHM procedure uses several swarm-based methods (a.k.a. inner metaheuristics, single metaheuristics, inner optimisers) as input and employs them in a hybrid mode for stage-based optimisation. The entire population is passed between inner metaheuristics without any losses. There is no limit to inner optimisers, although here we introduce the method based on five inner metaheuristics. This amount of inner cHM techniques is the result of a trade-off between efficiency (the fewer methods, the lower probing fit convergence) and optimisation performance (the more techniques, the higher hybridisation impact). 
All the inner metaheuristics need to be population-based, with a population composed of 
uniform individuals, so they are easily transferable between methods. For more sophisticated procedures, e.g. Bee Hive with different types of individuals (onlooker, employee bees, etc.), a population transfer strategy should be developed. The moment when the cHM algorithm passes population between single metaheuristics or algorithm phases is referred to as the \textit{switch point}.

cHM can be applied to solving any N-dimensional optimisation problem, but here we test it on standard benchmark functions in 2D space. Initially, the algorithm is populated with random individual solutions of a given size and shape. This population is then maintained in subsequent generations, and translated between the inner-optimisers selected by the cHM method. 

The algorithm alternates between two phases: \textit{probing} and \textit{fitting}, which are run in turns $n$ times or until the stopping criteria are met.

In a first stage---probing, the cHM algorithm iterates over all available inner methods and probes each of them for time $t_{probing}$ or $maxFE_{probing}$ (maximum number of function evaluations) to find the best one for a given optimisation phase. This stage is particularly important for the algorithm success, so the probing length should be carefully selected. Setting too small $t_{probing}$ or $maxFE_{probing}$ might result in sub-optimal inner metaheuristic selection, which affects all remaining training. On the other hand, too large values of $t_{probing}$ or $maxFE_{probing}$ may decrease the effect of various methods' hybridisation. In fact, it might cause an inaccurate metaheuristic switch point to be skipped and to lose the strength of different swarm-based methods.

In the second phase---fitting, cHM selects the best inner metaheuristics from the probing phase and performs the optimisation for time $t_{fit}$ or until $maxFE_{fit}$ is reached. After this process is finished, the population is saved, so it can be passed to all inner methods in the next probing stage. The population is passed to probing only if the value of the cost function is decreasing. Otherwise, the population from the previous probing stage is sent to the next probing phase without changes. This process is repeated for $n$ iterations or until the stopping criteria are met. It is worth noting that the fit phase performance is limited by the accuracy of the former stage. 

Thanks to this phase-based behaviour, the cHM method not only leverages the strengths of each inner metahueristic algorithm but also applies them in the optimal problem-solving stage. It is assumed that the probing stage estimates the best inner method for a given phase of the problem optimisation process. However, this feature is limited to the estimation error of the cHMs probing phase. Because of this, the cHM parameters should be precisely tuned.

The number of algorithm iterations ($n$) depends directly on the length of the probing and fit stages. Usually, it is assumed that the algorithm should converge in 3-5 iterations when  $maxFE_{probing}$ and  $maxFE_{fit}$ are aligned optimally. Naturally, for more complex problems, additional iterations may be required. Next, the probing and fit phase lengths should be selected jointly. Generally, to choose a nearly optimal value of these parameters, one should define a $probing-to-fit$ ratio as $maxFE_{probing}$ divided by $maxFE_{fit}$. This ratio defines what part of the algorithm should be spent on probing with regards to fit time. From our experiments, the probing-to-fit ratio should be between 0.2 and 0.5, depending on the optimised problem. Once the probing-to-fit ratio is established, the absolute values of the probing and fit stage lengths may be determined based on the problem's complexity.
A cHM pseudo-code is presented in the Algorithm \ref{algorithm:chm}.
\begin{algorithm}
	\caption{constrained Hybrid Mixed Metaheuristic optimisation} 
   
	\begin{algorithmic}[1]
    \State Assume $k$ metaheuristics with common characteristics $\theta_n$, that describe a population of solutions for a continuous optimisation problem, e.g. D-dimensional function $F$.
    The algorithm will be run on a maximum number of $F$ evaluations to ensure a reliable comparison of metaheuristics' performance.
    The cost function $C$ is used to compute the $F$ optimisation accuracy.
    \State Initialize each of $k$ metaheuristics with a random population $\theta_n$
    \For {$n=1,2,\ldots n$}
        \State Begin with metaheuristics probing
        \For {$k=1,2,\ldots k$}
            \While {$m_{probing}<MaxFE_{probing}$}
                \State Optimize function $F$ with metaheuristic $k$ and calculate $C$ value for each individual
            \EndWhile           
        \EndFor
        \State Select the best-performing $k-th$ metaheuristic and update $\theta_n$
        \While {$m_{fit}<MaxFE_{fit}$}
            \State Optimize function $F$ with metaheuristic $k$ using $\theta_n$ and compute $C$ value for each individual 
        \EndWhile
        \State Update $\theta_n$ 
        \If{$C$ value convergence for the function $F$ is met}
            \State Break
        \EndIf
	\EndFor
	\end{algorithmic}
    \label{algorithm:chm}
\end{algorithm}
%

\section{Experimental setup}
\label{sec:experimental_setup}

The performance of CHM was tested on 28 benchmark functions, listed in Table~\ref{table:function_definitions} in the Appendix. This performance was compared with the individual inner metaheuristic performance in Tables~\ref{table:mean_fitness_without_hpo} -~\ref{table:mean_fitness_inv_without_hpo}. 

For each benchmark function, the experiment was run 50 times. Computations were performed on a 12th Gen Intel Core i5-12450H 2.00 GHz with 16GB of RAM.
The fitness function was defined as an absolute difference between the function value for the proposed solution and the minimum function value:
%
\begin{equation}
    fitness = abs\left(F(\mathbf{X}) - F(\mathbf{X_0})\right),
    \label{equation:fitness}
\end{equation}
where $F$ represents a given benchmark function, \textbf{$X$} stands for a proposed solution and \textbf{$X_0$} is the optimal solution.
%
Also, the Euclidean distance between \textbf{$X$} and \textbf{$X_0$} was calculated to indicate the distance measure for the generated solution and the optimal one. ~\cite{euclidean}. 

To illustrate the performance characteristics of the optimisation methods, we aggregated the results using the \textit{mean}, \textit{sum}, \textit{minimum}, and \textit{standard deviation} statistics.

The following five inner metaherutistics were considered in this research:
\begin{itemize}
    \item Particle Swarm Optimisation (PSO) - a swarm-based optimisation technique grounded on the social behaviour of swarm animals or insects \cite{pso_origin}. PSO is widely used for different optimisation problems. In this method, the swarm of individuals is built from particles that, in their basic form, have position and movement velocity in the search space. During the algorithm iterations, the particles search the space of possible solutions, looking for the optimum. The particle properties are updated based on a fitness function. In addition, both cognitive (individual) and social (population) information is used to update each particle's behaviour. 
    \item Simulated Annealing (SA) - an algorithm following the annealing process of metallurgical materials, when they are cooled to low temperatures after heating~\cite{sa,aarts2013simulated}. This process is fairly grounded in physics, which brings a good statistical interpretability to the algorithm. The core characteristic of SA is the decrease in the probability of accepting worse solutions as the algorithm iterates more (decreasing temperature). This process allows the method to converge effectively and correctly steers it to escape local minima. SA is especially powerful for combinatorial problems.
    \item Genetic Algorithm (GA) - inspired by Darwinian natural selection principles~\cite{ga,alexakis2025genetic}. In this procedure, the population of individuals evolves through genetic operations: crossover, mutation, and selection. Individuals are assessed by their fitness value, and fitter individuals pass to the next generation (algorithm iteration) with higher probability. GA is widely used for complex, multimodal, and nonlinear problems.
    \item Differential Evolution (DE) - an evolutionary algorithm that uses different from GA genetic operations to find the best solution over a swarm of individuals \cite{de,das2016recent}. Unlike GA, DE uses the scaled vector difference of individuals' fitnesses to perform population evolution, rather than randomised or linear operations. This makes the selection process more robust for continuous optimisation problems.
    \item Bacterial Foraging Optimisation (BFO) - a modern, swarm-based optimisation method inspired by the foraging behaviour of E. coli bacteria \cite{bfo,passino2012bacterial}. The method models different bacteria swarming mechanisms to perform optimisation:  chemotaxis (movement), swarming, reproduction, and elimination-dispersal. These different search methods are responsible for local and global search in the optimisation space, as well as for random movements. BFO is especially well-suited for complex and noisy problems.
\end{itemize}

The cHM's inner metaheuristic selection is based on the ease of population transfer between methods. In addition, the selected methods consist of state-of-the-art swarm methods like PSO and SA, well-grounded evolutionary methods: DE, GA, and more recent techniques such as BFO. It is assumed that this combination could allow the cHM algorithm to take advantage of the strengths of different types of optimisation techniques.

The following parameters were used in the cHM implementation for all experiments: $n=4$, $population\_size=20$. As a consequence of the varying characteristics of different benchmark functions, $maxFE_{probing}=100$, $maxFE_{fit}=200$ parameters were set accordingly to the function type. The function buckets are presented in Table \ref{tab:function_buckets_parameters}. Notably, the function grouping is subjective, based on the author's scientific experience. The values of  $maxFE_{probing}=100$, $maxFE_{fit}=200$ were established within the regular HPO procedure for these two cHM arguments.

\begin{figure}
    \centering
    \includegraphics[width=0.9\linewidth]{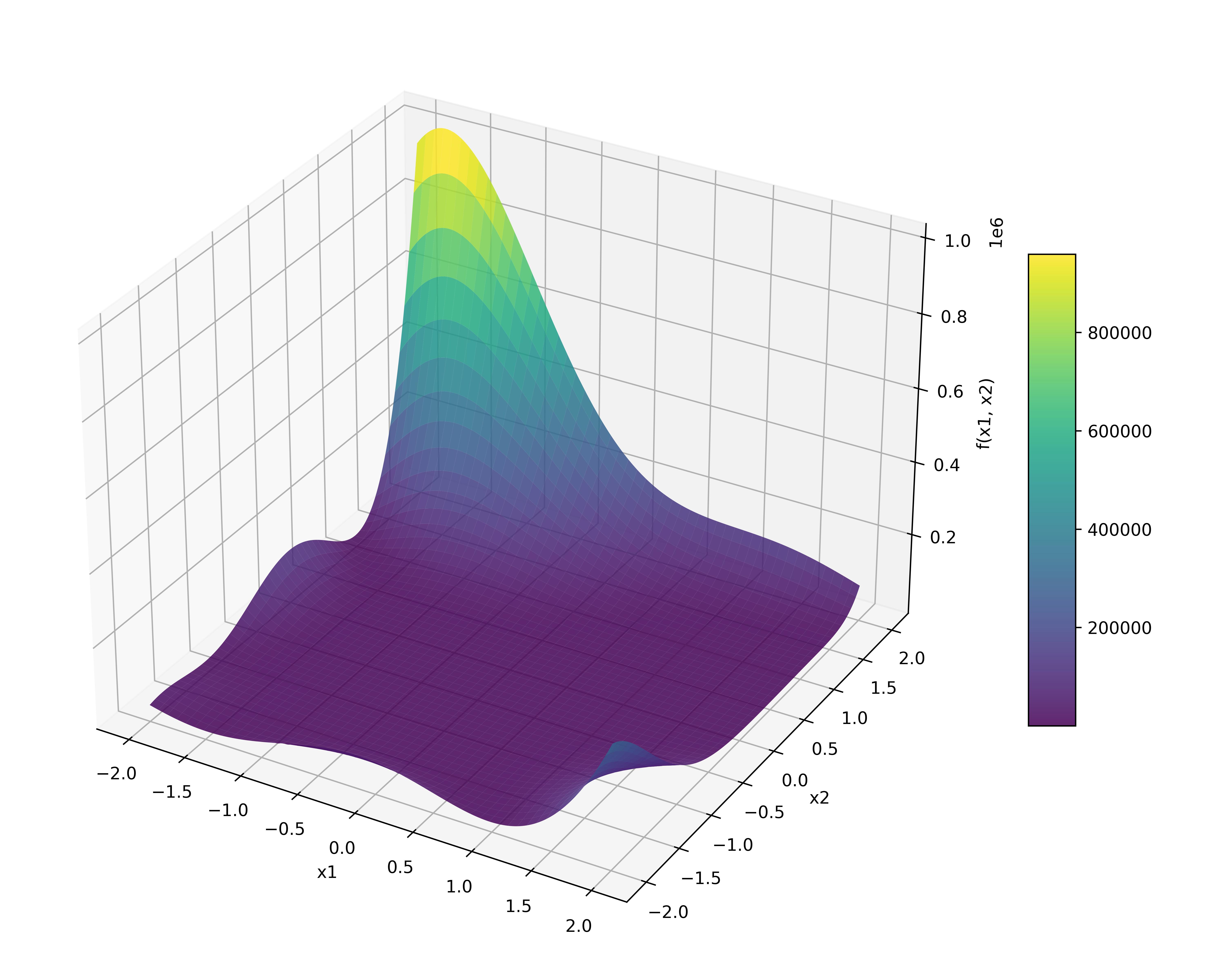}
    \caption{Goldstein-Price function illustration for range $x \epsilon [-2;2]$.}
    \label{fig:Goldstein-Price}
\end{figure}

We tested cHM on 28 well-known optimisation problems (\url{https://infinity77.net/global_optimisation/}) in a 2D space. 
These functions represent classic problems used in optimisation heuristic benchmarking, e.g., in the CEC competitions. A full description of all benchmark functions is available in the Appendix (Table~\ref{table:function_definitions} and Figure~\ref{fig:benchmark_gallery_full}). One example is the Goldstein-Price function plotted in Figure \ref{fig:Goldstein-Price} for $x \epsilon [-2;2]$ and defined by the equation: 


\begin{align*}\label{equation:exemplary_function}
f(x) = &\left(1 + \left(x_1 + x_2 + 1\right)^2 \left(19 - 14x_1 + 3x_1^2 - 14x_2 + 6x_1x_2 + 3x_2^2\right)\right) \\
       &\cdot \left(30 + \left(2x_1 - 3x_2\right)^2 \left(18 - 32x_1 + 12x_1^2 + 48x_2 - 36x_1x_2 + 27x_2^2\right)\right).
\end{align*}

%
%

\section{Results}
\label{sec:results}

\subsection{General results}
\label{subsec:general}

In Tables~\ref{table:mean_fitness_without_hpo} -~\ref{table:min_fitness_without_hpo}, we present aggregated comparisons of 50 runs between optimisation with the cHM and its inner metaheuristics. The results are rounded to 3 decimal places, and when the fitness value is lower than 1e-04, the results are presented as $0.0$. Consequently, there is no marking for the lowest fitness value in these tables as it would require more decimal places, which would make the data illegible. On the whole, it can be seen that DE and cHM achieve the lowest fitness-related scores across various aggregations. Furthermore, several functions have significantly lower performance than others, especially for BFO and SA. For example, the \textit{Goldstein-Price} and \textit{Quadratic} functions have relatively higher values of the aggregated mean fitness than other functions. It may be explained by exponential components in their formulae, which can cause a function value explosion during optimisation. 

Table~\ref{table:mean_fitness_without_hpo} describes the mean best fitness values over 50 experiment runs. In general, the DE, GA and cHM methods performed similarly and gave the best results for all functions. It may be noticed that for the \textit{Bird} function, DE was even slightly better than cHM (when comparing the 3rd decimal places). Likewise, the GA metaheuristic performed a little better than cHM for the \textit{Quadratic} function. It may arise from the overhead of the probing phase and could be mitigated through the CHM's parameter tuning for these functions. On the other hand, for all other functions, cHM achieved the best performance, clearly producing the lowest mean best fitness value overall. Indeed, the final mean fitness is lower for cHM than for the best algorithm among the single optimisers for 23 out of 28 functions. It showcases the hybridisation strength of the proposed method and confirms that it generally selects optimal inner-optimisers over the algorithm phases. Additionally, for the functions where cHM yields worse results, the absolute difference between the best method and cHM is usually negligibly small.

Furthermore, metaheuristics' performance in Table \ref{table:mean_fitness_without_hpo} exhibits variability for different benchmark buckets (demonstrated in Table \ref{tab:function_buckets_parameters}). To exemplify, for the Single-basin functions, the mean fitness value was approximately near 0, except for the \textit{Quadratic} function, where only cHM and GA showed relatively low fitness function value among all tested procedures. By contrast, the Ill-conditioned benchmarks didn't exhibit a common characteristic across the entire bucket. In fact, for the \textit{Rosenbrock} benchmark, mean fitness function values were around 1, while for the \textit{Brown}, the loss values were equal to 0 for almost all methods. Afterwards, the highly-multimodal generally performed similarly for all functions, with the fitness function values near 0. As this group contains many functions, there were examples of perfect optimisation for (\textit{Keane}, \textit{Levy03}) functions and slightly worse optimisation for (\textit{Rastrigin}, \textit{Bochanevsky01}) functions. On the other hand, the functions from the Few well-separate global minima bucket gave the worst general results among all buckets. In particular, most metaheuristic methods yielded near-zero mean fitness values only for the \textit{Zettl}, \textit{Hosaki}, and \textit{Treccani} functions. In the end, the Shifted minima group performed well in general, with the mean fitness loss value around 0.


As shown in Table~\ref{table:std_fitness_without_hpo}, the standard deviation (std) of the best mean fitness values across 50 runs is high, of the same magnitude as the mean fitness value. This is mainly caused by the sensitivity of the optimisation process to the initialisation of the population. 
Generally, the distribution of the std values across the benchmarks is similar to the one shown in Table~\ref{table:mean_fitness_without_hpo}. DE and cHM generated the lowest values of std; however, the results depend on a particular benchmark. Conversely, SA showed the highest standard deviation. In fact, mean fitness values for different functions yield similar results to the std of the fitness function, which is an expected behaviour for heuristic-like systems. 
High \textit{std} value suggests that when applying population-based methods, the experiment should be run multiple times to generate reliable results, which is a common observation in this kind of research. Alternatively, experimental runs exhibiting exceptionally high mean and standard deviation of fitness values could be excluded from the aggregated results as potentially invalid, for instance, due to suboptimal population initialisation. Nevertheless, all runs were retained in this study to ensure data transparency and completeness.

The minimum fitness value of cHM across 50 runs 
(Table~\ref{table:min_fitness_without_hpo}) 
equals zero (a perfect fitness) for all 28 benchmark functions.
DE and GA 
demonstrate comparable performance. 
However, such consistency is expected given the number of repetitions performed. 
Overall, the cHM performance on 28 benchmarks outperforms that of any individual optimisers, with DE and GA being superior among individual metaheuristics, with the mean fitness values often similar to cHM.

\begin{table}[]
\caption{Mean best fitness values across 50 experiment runs for different optimisation methods.}
\begin{tabular}{|c|c|c|c|c|c|c|}
\hline
\textit{\textbf{Function name}} & \textbf{BFO} & \textbf{DE} & \textbf{GA} & \textbf{PSO} & \textbf{SA} & \textbf{cHM} \\ \hline
\textit{Ackley02}         & 0.323 & 0.119 & 0.013 & 0.198 & 24.93   & 0.001 \\ \hline
\textit{Beale}            & 0.103 & 0.088 & 0.097 & 0.027 & 3.778   & 0.001 \\ \hline
\textit{Bird}             & 4.594 & 0.013 & 1.551 & 2.167 & 53.89  & 0.163 \\ \hline
\textit{Bohachevsky01}    & 0.736 & 0.052 & 0.000 & 0.049 & 17.08  & 0.000   \\ \hline
\textit{Branin02}         & 1.552 & 0.923 & 1.637 & 1.228 & 6.040    & 0.676 \\ \hline
\textit{Brent}            & 0.006 & 0.656 & 0.000   & 0.094 & 21.76  & 0.000   \\ \hline
\textit{Brown}            & 0.006 & 0.000   & 0.000   & 0.008 & 0.252   & 0.000   \\ \hline
\textit{Eggcrate}         & 4.714 & 0.000   & 0.120  & 0.076 & 12.83  & 0.000   \\ \hline
\textit{Goldstein-Price}   & 5.875 & 0.544 & 0.011 & 8.271 & 91.01  & 0.003 \\ \hline
\textit{Himmelblau}       & 0.178 & 0.001 & 0.003 & 0.189 & 17.88  & 0.002 \\ \hline
\textit{Hosaki}           & 0.095 & 0.000   & 0.001 & 0.073 & 2.991   & 0.001 \\ \hline
\textit{Keane}            & 0.000   & 0.000   & 0.000   & 0.000   & 0.000     & 0.000   \\ \hline
\textit{Levy03}           & 0.000   & 0.000   & 0.000   & 0.000   & 0.355   & 0.000   \\ \hline
\textit{Matyas}           & 0.004 & 0.003 & 0.000   & 0.001 & 0.630    & 0.000   \\ \hline
\textit{Price02}          & 0.099 & 0.045 & 0.048 & 0.054 & 0.233   & 0.005 \\ \hline
\textit{Quadratic}        & 0.796 & 4.244 & 0.007 & 0.758 & 842.3 & 0.041 \\ \hline
\textit{Rastrigin}        & 10.54 & 0.000   & 0.003 & 0.060  & 20.05   & 0.000   \\ \hline
\textit{Rosenbrock}       & 1.263 & 0.508 & 0.123 & 0.174 & 236.1 & 0.010  \\ \hline
\textit{Rotatedellipse01} & 0.048 & 0.005 & 0.000   & 0.060  & 48.92  & 0.001 \\ \hline
\textit{Salomon}          & 0.462 & 0.000   & 0.001 & 0.002 & 0.929   & 0.000   \\ \hline
\textit{Schaffer03}       & 0.491 & 0.003 & 0.097 & 0.001 & 0.485   & 0.000   \\ \hline
\textit{Schaffer04}       & 0.031 & 0.000   & 0.002 & 0.000   & 0.035   & 0.000   \\ \hline
\textit{Schwefel04}       & 0.042 & 0.040  & 0.032 & 0.010  & 2.113   & 0.000   \\ \hline
\textit{Treccani}         & 0.011 & 0.000   & 0.000   & 0.008 & 1.539   & 0.000   \\ \hline
\textit{Ursem04}          & 0.050  & 0.000   & 0.002 & 0.002 & 0.456   & 0.000   \\ \hline
\textit{Whitley}          & 0.072 & 0.003 & 0.001 & 0.007 & 187.1 & 0.001 \\ \hline
\textit{Zettl}            & 0.010  & 0.002 & 0.000   & 0.006 & 0.285   & 0.000   \\ \hline
\textit{Zirilli}          & 0.076 & 0.004 & 0.002 & 0.014 & 3.796   & 0.000   \\ \hline
\end{tabular}
\label{table:mean_fitness_without_hpo}
\end{table}
\begin{table}[]
\caption{Standard deviations of the mean best fitness values across 50 experiment runs for different optimisation methods.}
\begin{tabular}{|c|c|c|c|c|c|c|}
\hline
\textit{\textbf{Function name}} & \textbf{BFO} & \textbf{DE} & \textbf{GA} & \textbf{PSO} & \textbf{SA} & \textbf{cHM} \\ \hline
\textit{Ackley02}         & 0.174  & 0.724  & 0.016 & 0.25  & 15.14  & 0.005 \\ \hline
\textit{Beale}            & 0.287  & 0.226  & 0.258 & 0.038 & 3.890    & 0.004 \\ \hline
\textit{Bird}             & 10.53 & 0.088  & 4.816 & 3.143 & 29.83  & 0.600   \\ \hline
\textit{Bohachevsky01}    & 0.387  & 0.248  & 0.001 & 0.116 & 15.33  & 0.000   \\ \hline
\textit{Branin02}         & 1.458  & 0.791  & 0.643 & 0.404 & 3.443   & 0.735 \\ \hline
\textit{Brent}            & 0.003  & 2.512  & 0.000   & 0.088 & 17.59  & 0.000   \\ \hline
\textit{Brown}            & 0.003  & 0.001  & 0.000   & 0.013 & 0.154   & 0.000   \\ \hline
\textit{Eggcrate}         & 5.487  & 0.000    & 0.479 & 0.138 & 5.168   & 0.000   \\ \hline
\textit{Goldstein-Price}   & 10.19  & 3.818  & 0.016 & 8.141 & 102.2 & 0.017 \\ \hline
\textit{Himmelblau}       & 0.139  & 0.003  & 0.009 & 0.218 & 17.55  & 0.008 \\ \hline
\textit{Hosaki}           & 0.099  & 0.000    & 0.002 & 0.076 & 3.073   & 0.003 \\ \hline
\textit{Keane}            & 0.000    & 0.000    & 0.000   & 0.000   & 0.000     & 0.000   \\ \hline
\textit{Levy03}           & 0.000    & 0.000    & 0.000   & 0.000   & 0.468   & 0.000   \\ \hline
\textit{Matyas}           & 0.016  & 0.020   & 0.000   & 0.001 & 0.686   & 0.000   \\ \hline
\textit{Price02}          & 0.023  & 0.049  & 0.043 & 0.047 & 0.124   & 0.020  \\ \hline
\textit{Quadratic}        & 0.512  & 19.79 & 0.012 & 0.935 & 861.2 & 0.119 \\ \hline
\textit{Rastrigin}        & 14.67 & 0.003  & 0.006 & 0.101 & 22.85  & 0.001 \\ \hline
\textit{Rosenbrock}       & 1.830   & 1.201  & 0.320  & 0.349 & 618.4 & 0.021 \\ \hline
\textit{Rotatedellipse01} & 0.039  & 0.029  & 0.001 & 0.124 & 53.89  & 0.004 \\ \hline
\textit{Salomon}          & 0.541  & 0.000    & 0.001 & 0.003 & 0.653   & 0.000   \\ \hline
\textit{Schaffer03}       & 0.036  & 0.021  & 0.177 & 0.002 & 0.048   & 0.001 \\ \hline
\textit{Schaffer04}       & 0.014  & 0.000    & 0.007 & 0.000   & 0.011   & 0.000   \\ \hline
\textit{Schwefel04}       & 0.227  & 0.181  & 0.136 & 0.009 & 2.556   & 0.000   \\ \hline
\textit{Treccani}         & 0.006  & 0.000    & 0.000   & 0.011 & 1.805   & 0.000   \\ \hline
\textit{Ursem04}          & 0.027  & 0.000    & 0.001 & 0.003 & 0.209   & 0.000   \\ \hline
\textit{Whitley}          & 0.093  & 0.014  & 0.002 & 0.024 & 510.6 & 0.002 \\ \hline
\textit{Zettl}            & 0.043  & 0.009  & 0.000   & 0.007 & 0.238   & 0.001 \\ \hline
\textit{Zirilli}          & 0.097  & 0.028  & 0.007 & 0.015 & 5.010    & 0.000   \\ \hline
\end{tabular}
\label{table:std_fitness_without_hpo}
\end{table}
\begin{table}[]
\caption{Minimum best fitness values across 50 experiment runs for different optimisation methods.}
\begin{tabular}{|c|c|c|c|c|c|c|}
\hline
\textit{\textbf{Function name}} & \textbf{BFO} & \textbf{DE} & \textbf{GA} & \textbf{PSO} & \textbf{SA} & \textbf{cHM} \\ \hline
\textit{Ackley02}         & 0.059 & 0.000 & 0.001 & 0.000   & 3.951  & 0.000 \\ \hline
\textit{Beale}            & 0.000   & 0.000 & 0.000   & 0.000   & 0.082  & 0.000 \\ \hline
\textit{Bird}             & 0.006 & 0.000 & 0.000   & 0.028 & 4.239  & 0.000 \\ \hline
\textit{Bohachevsky01}    & 0.037 & 0.000 & 0.000   & 0.000   & 0.566  & 0.000 \\ \hline
\textit{Branin02}         & 0.007 & 0.000 & 0.000   & 0.110  & 0.663  & 0.000 \\ \hline
\textit{Brent}            & 0.000   & 0.000 & 0.000   & 0.002 & 0.104  & 0.000 \\ \hline
\textit{Brown}            & 0.001 & 0.000 & 0.000   & 0.000   & 0.005  & 0.000 \\ \hline
\textit{Eggcrate}         & 0.015 & 0.000 & 0.000   & 0.000   & 1.610   & 0.000 \\ \hline
\textit{Goldstein-Price}   & 0.084 & 0.000 & 0.000   & 0.012 & 0.699  & 0.000 \\ \hline
\textit{Himmelblau}       & 0.000   & 0.000 & 0.000   & 0.000   & 0.106  & 0.000 \\ \hline
\textit{Hosaki}           & 0.002 & 0.000 & 0.000   & 0.001 & 0.002  & 0.000 \\ \hline
\textit{Keane}            & 0.000   & 0.000 & 0.000   & 0.000   & 0.000    & 0.000 \\ \hline
\textit{Levy03}           & 0.000   & 0.000 & 0.000   & 0.000   & 0.000    & 0.000 \\ \hline
\textit{Matyas}           & 0.000   & 0.000 & 0.000   & 0.000   & 0.014  & 0.000 \\ \hline
\textit{Price02}          & 0.007 & 0.000 & 0.000   & 0.000   & 0.029  & 0.000 \\ \hline
\textit{Quadratic}        & 0.004 & 0.000 & 0.000   & 0.000   & 21.20 & 0.000 \\ \hline
\textit{Rastrigin}        & 0.026 & 0.000 & 0.000   & 0.000   & 0.056  & 0.000 \\ \hline
\textit{Rosenbrock}       & 0.001 & 0.000 & 0.000   & 0.006 & 0.114  & 0.000 \\ \hline
\textit{Rotatedellipse01} & 0.000   & 0.000 & 0.000   & 0.000   & 0.674  & 0.000 \\ \hline
\textit{Salomon}          & 0.003 & 0.000 & 0.000   & 0.000   & 0.057  & 0.000 \\ \hline
\textit{Schaffer03}       & 0.267 & 0.000 & 0.000   & 0.000   & 0.174  & 0.000 \\ \hline
\textit{Schaffer04}       & 0.000   & 0.000 & 0.000   & 0.000   & 0.000    & 0.000 \\ \hline
\textit{Schwefel04}       & 0.000   & 0.000 & 0.000   & 0.000   & 0.006  & 0.000 \\ \hline
\textit{Treccani}         & 0.000   & 0.000 & 0.000   & 0.000   & 0.006  & 0.000 \\ \hline
\textit{Ursem04}          & 0.000   & 0.000 & 0.000   & 0.000   & 0.001  & 0.000 \\ \hline
\textit{Whitley}          & 0.000   & 0.000 & 0.000   & 0.000   & 0.009  & 0.000 \\ \hline
\textit{Zettl}            & 0.000   & 0.000 & 0.000   & 0.000   & 0.001  & 0.000 \\ \hline
\textit{Zirilli}          & 0.000   & 0.000 & 0.000   & 0.000   & 0.232  & 0.000 \\ \hline
\end{tabular}
\label{table:min_fitness_without_hpo}
\end{table}

\subsection{cHM vs inner optimisers}

In order to determine whether the inner-optimiser selection in the probing phase is efficient, the cHM performance was compared with each individual metaheuristic (solving a given problem). Table~\ref{table:iteration_chm_single_merged} presents metaheuristics that yielded the lowest mean fitness function versus the ones that were most often selected by the cHM. In general, DE appeared to be superior among individual metaheuristics. At the same time, it was most often chosen by cHM. Notably, the values in the \textit{Single best} and \textit{Best cHM} columns often differ, indicating that the cHM framework did not always select the metaheuristic that achieved the best performance in the single-method runs. This discrepancy may be attributed to the phase-like nature of cHM, which dynamically selects the most suitable algorithm for each stage of the optimisation process. For instance, the DE algorithm often exhibits high accuracy during the initial phases of benchmark optimisation, which may explain its frequent selection by cHM. For benchmarks where DE was not the best individual method, cHM tended to favor the same inner optimisers for the \textit{Keane}, \textit{Schaffer03}, and \textit{Rosenbrock} benchmarks. Conversely, for benchmarks such as \textit{Schwefel04}, \textit{Quadratic}, and \textit{Matyas}, cHM selected alternative methods instead of the best single optimiser.

To further analyse the relation between metaheuristics and the cHM probing procedure, Table~\ref{table:iteration_chm_single_merged} presents the number of times each metaheuristic was selected by cHM versus its individual performance. 
Generally, DE and PSO were selected most often by cHM, followed by GA. These results are consistent with the strength of single metaheuristics, as DE, PSO, and GA are the only three functions that yielded the lowest fitness scores. For instance, for the \textit{Rosenbrock} and \textit{Schaffer03} functions, the best single metaheuristics were the same as the most often selected inner optimiser in the cHM algorithm. On the other hand, \textit{Schwefel04} and \textit{Matyas} are examples of incompatibility between cHM and the best single optimiser. In contrast, BFO and SA were selected the fewest times among all metaheuristics. This observation is consistent with the data presented in Tables \ref{table:mean_fitness_without_hpo}–\ref{table:std_fitness_without_hpo}, as, overall, BFO and SA produced the highest aggregated fitness values across all methods.

\begin{table}[htbp]
\centering
\caption{Best individual metaheuristics versus the frequency of metaheuristic selection by the cHM algorithm. The metric is the mean best fitness value over 50 runs.}
\renewcommand{\arraystretch}{1.1}
\begin{tabular}{|l|c|c|c|c|c|c|c|}
\hline
\textbf{Function name} & \textbf{Single best} & \textbf{Best cHM} & \textbf{hmm\_DE} & \textbf{hmm\_PSO} & \textbf{hmm\_BFO} & \textbf{hmm\_GA} & \textbf{hmm\_SA} \\ \hline
\textit{Ackley02}         & GA  & DE  & 64 & 28 & 0  & 5  & 2  \\ \hline
\textit{Beale}            & PSO & DE  & 67 & 39 & 23 & 24 & 18 \\ \hline
\textit{Bird}             & DE & DE  & 73 & 19 & 10 & 14 & 12 \\ \hline
\textit{Bohachevsky01}    & GA  & DE  & 37 & 16 & 2  & 5  & 1  \\ \hline
\textit{Branin02}         & DE  & DE  & 100 & 20 & 25 & 36 & 11 \\ \hline
\textit{Brent}            & GA  & DE  & 56 & 9  & 1  & 9  & 3  \\ \hline
\textit{Brown}            & DE/GA  & DE  & 50 & 2  & 0  & 1  & 4  \\ \hline
\textit{Eggcrate}         & DE  & DE  & 42 & 12 & 0  & 5  & 0  \\ \hline
\textit{Goldstein-Price}   & GA  & DE  & 78 & 2  & 10 & 24 & 14 \\ \hline
\textit{Himmelblau}       & DE  & DE  & 81 & 8  & 12 & 21 & 21 \\ \hline
\textit{Hosaki}           & DE  & SA  & 48 & 31 & 20 & 42 & 59 \\ \hline
\textit{Keane}            & ALL  & PSO & 10 & 21 & 8  & 14 & 4  \\ \hline
\textit{Levy03}           & \begin{tabular}[c]{@{}c@{}}BFO / DE \\ GA / PSO\end{tabular}  & DE  & 40 & 2  & 1  & 9  & 2  \\ \hline
\textit{Matyas}           & GA  & PSO & 40 & 47 & 10 & 22 & 17 \\ \hline
\textit{Price02}          & DE  & PSO & 35 & 40 & 7  & 26 & 13 \\ \hline
\textit{Quadratic}        & GA  & DE  & 78 & 31 & 9  & 20 & 13 \\ \hline
\textit{Rastrigin}        & DE & DE  & 56 & 29 & 7  & 9  & 5  \\ \hline
\textit{Rosenbrock}       & GA  & GA  & 48 & 32 & 21 & 58 & 39 \\ \hline
\textit{Rotatedellipse01} & GA  & DE  & 55 & 32 & 8  & 21 & 10 \\ \hline
\textit{Salomon}          & DE  & DE  & 46 & 23 & 0  & 3  & 3  \\ \hline
\textit{Schaffer03}       & PSO  & PSO  & 52 & 70 & 16 & 22 & 31 \\ \hline
\textit{Schaffer04}       & DE/PSO  & DE  & 50 & 41 & 18 & 37 & 54 \\ \hline
\textit{Schwefel04}       & PSO & DE  & 82 & 12 & 9  & 10 & 6  \\ \hline
\textit{Treccani}         & DE/GA  & DE  & 54 & 10 & 1  & 8  & 1  \\ \hline
\textit{Ursem04}          & DE  & DE  & 48 & 32 & 1  & 3  & 2  \\ \hline
\textit{Whitley}          & GA  & DE  & 52 & 24 & 5  & 33 & 35 \\ \hline
\textit{Zettl}            & GA & DE  & 60 & 11 & 11 & 13 & 8  \\ \hline
\textit{Zirilli} & GA & DE & 52 & 2 & 3 & 11 & 2 \\ \hline

\end{tabular}
\label{table:iteration_chm_single_merged}
\end{table}

\subsection{Aggregated results}

Since the performance of the population-based methods differs between benchmark functions, the aggregated performance results 
are shown in Table~\ref{table:aggregated_euclidean_fitness_distance}.
The left part of the table presents the count of the methods with the lowest mean best value for all benchmark functions over 50 experiment runs. The cHM algorithm presents the best performance for 17 benchmarks. The second-best method is DE, whose results are aligned with the results for each particular function presented in the previous sections. In addition, the GA method gave the lowest mean fitness score for 5 functions, which corresponds to the results shown in Tables \ref{table:mean_fitness_without_hpo} - \ref{table:iteration_chm_single_merged}.
 
Similarly to the fitness value, cHM achieved the lowest Euclidean distance most frequently, for 16 of 28 benchmark functions. 
There is no clear second-best method, as PSO,
and GA achieved similar aggregated scores (6). Generally speaking, these results are consistent with the data for individual benchmarks shown in Tables \ref{table:mean_fitness_without_hpo} - \ref{table:iteration_chm_single_merged}

In the right part of Table~\ref{table:aggregated_euclidean_fitness_distance}, the average mean and the sum of the mean of the best fitness values are presented for all the benchmarks across 50 repetitions of the experiments for all 6 optimisation methods. Overall, cHM shows the lowest values in terms of both average and sum aggregations. The second-best methods are the DE and GA methods, with clearly lower average fitness results than the remaining methods.

In general, PSO and BFO produced similar results. The outstanding performance of DE and GA compared to other metaheuristics is mainly caused by the differential nature of these methods, which is especially well-suited for continuous optimisation problems. 
Conversely, the notably high aggregated fitness values observed for the SA methods may be attributed to the inherent characteristics of the algorithm. Specifically, SA does not effectively exploit the collective memory of the entire population of solutions, and its exploration–exploitation balance is challenging to control due to the simplicity of its recombination mechanism.

\begin{table}[htbp]
\centering
\caption{Summary of results for all 28 benchmark functions across 50 experiment repetitions. 
Left: number of times each method achieved the lowest mean best fitness and Euclidean distance. 
Right: aggregated average and sum of the mean best fitness values.}
\renewcommand{\arraystretch}{1.1}
\begin{tabular}{|c|cc|cc|}
\hline
\textbf{Method} 
& \textbf{Fitness Function} 
& \textbf{Euclidean Distance} 
& \textbf{Average Fitness} 
& \textbf{Sum of Fitness} \\ \hline

\textit{cHM} & \textbf{17} & \textbf{16} & \textbf{0.032} & \textbf{0.905} \\ \hline
\textit{DE}  & 6 & 0 & 0.259 & 7.252 \\ \hline
\textit{PSO} & 0 & 6 & 0.484 & 13.54 \\ \hline
\textit{GA}  & 5 & 6 & 0.134 & 3.751 \\ \hline
\textit{BFO} & 0 & 0 & 1.149 & 21.17 \\ \hline
\textit{SA}  & 0 & 0 & 57.060 & 1598 \\ \hline

\end{tabular}
\label{table:aggregated_euclidean_fitness_distance}
\end{table}

\subsection{Iterations}
\label{subsec:iterations}

Table \ref{table:mean_fitness_inv_without_hpo} shows the mean number of invocations of the fitness function for different heuristic methods for each optmisation benchmark. In general, cHM required a medium number of fitness function calls.

This shows that the hybrid algorithm has some overhead caused by the probing phase, but it is still capable of computing strong solutions in a reasonable time. Moreover, cHM can integrate various methods without domain-specific knowledge, which can help to skip the optimisation technique selection process.

\begin{figure}
    \centering
    \subfigure{\includegraphics[width=0.9\textwidth]{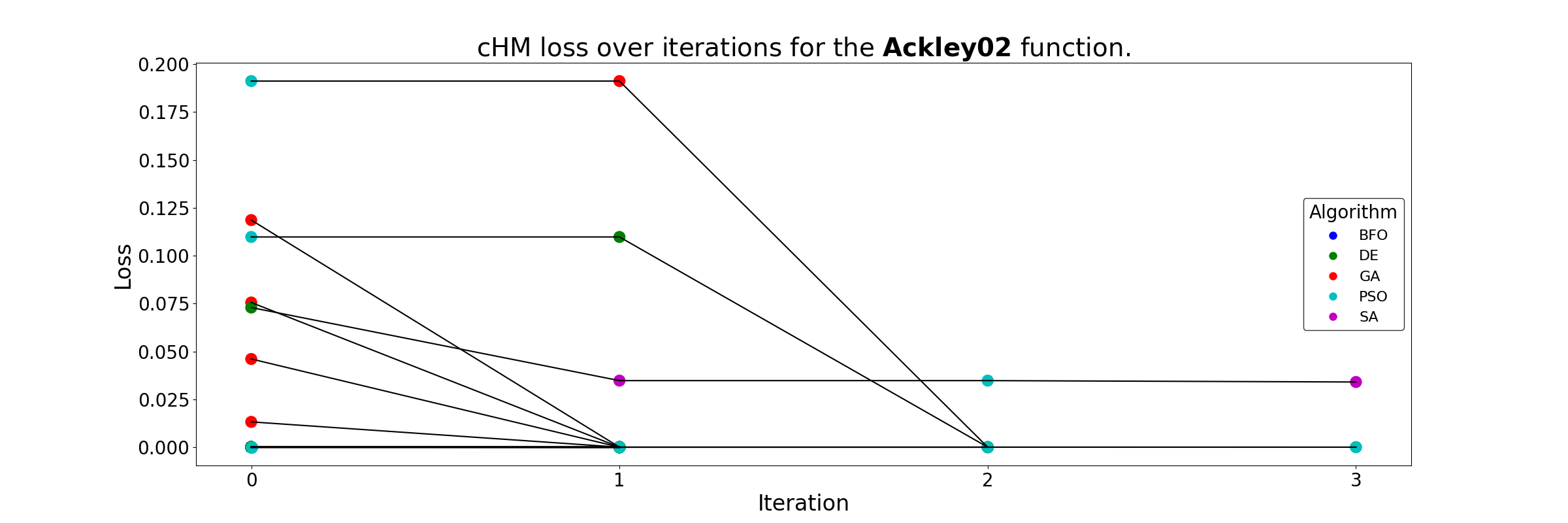}} 
    \subfigure{\includegraphics[width=0.9\textwidth]{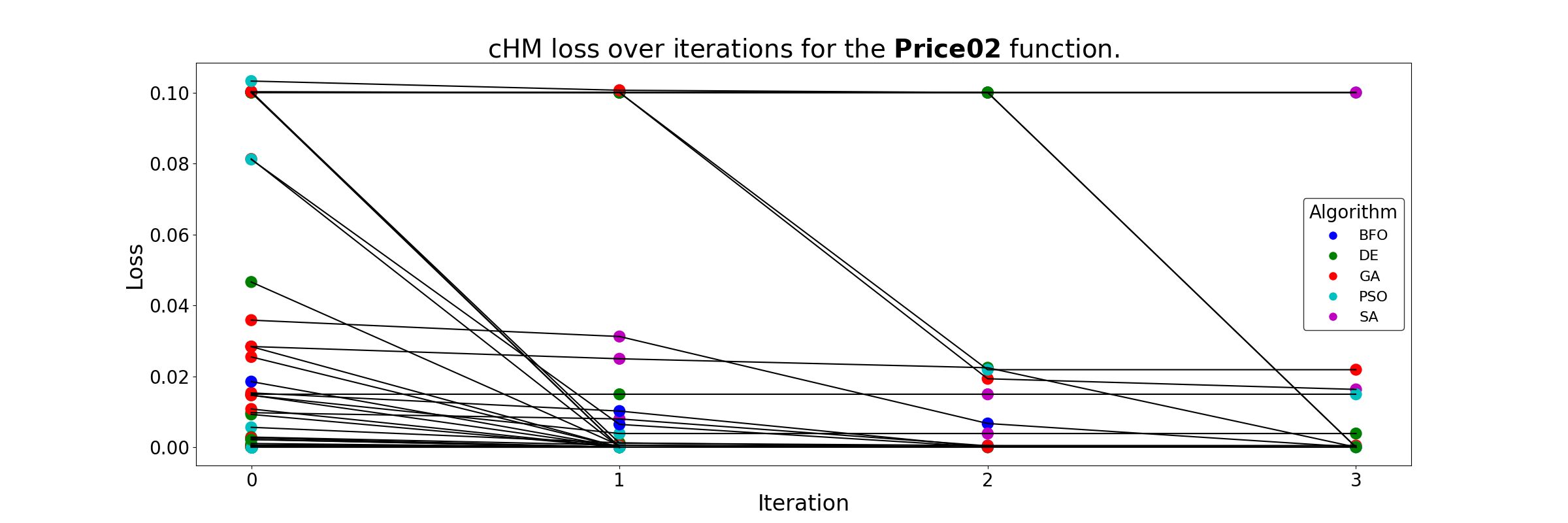}} 
    \subfigure{\includegraphics[width=0.9\textwidth]{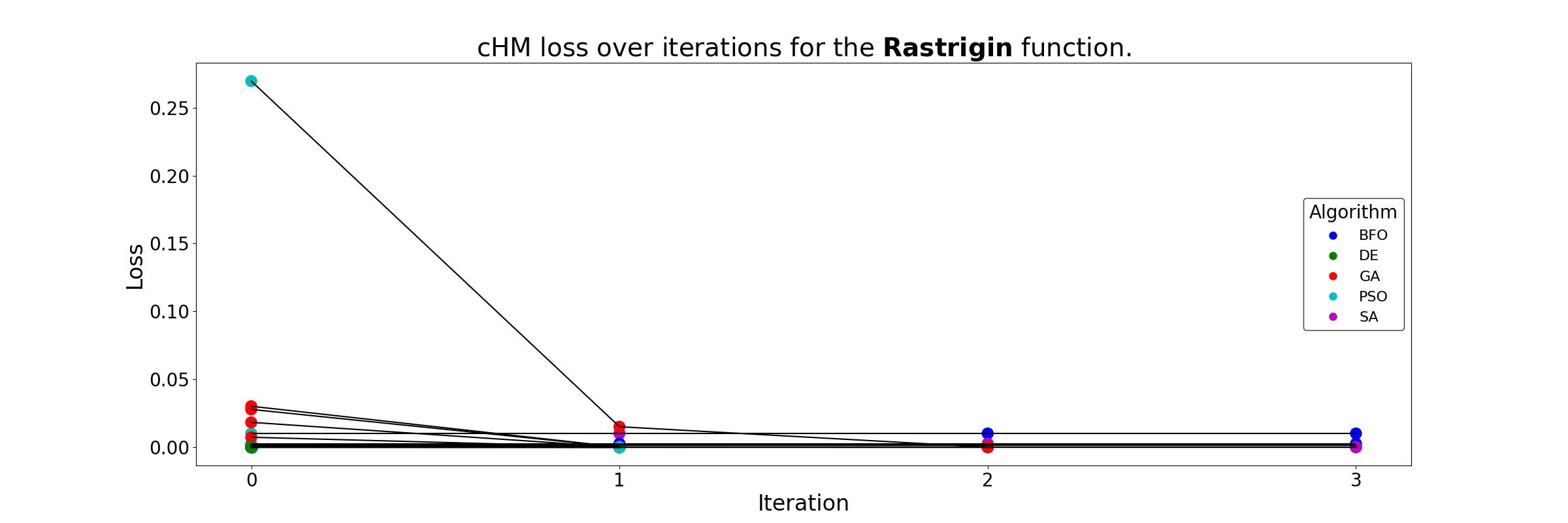}} 
    \caption{Illustration of the cHM optimisation behaviour over iterations for different benchmark functions (\textit{Ackley02}, \textit{Price02} and \textit{Rastrigin}, respectively. Each figure plots 50 experimental runs).}
    \label{fig:iterations}
\end{figure}

Generally, the BFO method needed the highest number of fitness function calls to reach the stopping criteria, which is expected as it is the most complex technique among all single-optimisers. Conversely, DE exhibited the lowest number of function invocations among all evaluated methods, which, combined with its overall fitness performance, positions it as the primary competitor to the cHM algorithm.
\begin{table}[]
\caption{Average values of fitness invocations for different optimisation methods for 50 experiment runs. Methods with the highest and lowest values for each function are marked in bold.}
\begin{tabular}{|c|c|c|c|c|c|c|}
\hline
\textit{\textbf{Function name}} & \textbf{DE} & \textbf{GA} & \textbf{SA} & \textbf{PSO} & \textbf{BFO} & \textbf{cHM} \\ \hline
\textit{Ackley02}         & \textbf{283} & 546          & 434          & 479          & 822 & 394          \\ \hline
\textit{Beale}            & 447          & 446          & 411          & 333          & 616 & \textbf{307} \\ \hline
\textit{Bird}             & \textbf{211} & 362          & 344          & 327          & 621 & 213          \\ \hline
\textit{Bohachevsky01}    & 551          & 490          & 434          & \textbf{347} & 819 & 573          \\ \hline
\textit{Branin02}         & 468          & 410          & 383          & \textbf{357} & 622 & 457          \\ \hline
\textit{Brent}            & 478          & 498          & 466          & 467          & 822 & \textbf{346} \\ \hline
\textit{Brown}            & 382          & 380          & 366          & \textbf{261} & 616 & 375          \\ \hline
\textit{Eggcrate}         & 466          & \textbf{458} & \textbf{458} & 487          & 823 & 473          \\ \hline
\textit{Goldstein-Price}   & \textbf{158} & 380          & 350          & 339          & 608 & 310          \\ \hline
\textit{Himmelblau}       & 248          & 392          & 366          & 339          & 621 & \textbf{247} \\ \hline
\textit{Hosaki}           & 416          & \textbf{350} & 366          & 357          & 599 & 402          \\ \hline
\textit{Keane}            & 244          & \textbf{150} & 388          & 319          & 551 & 177          \\ \hline
\textit{Levy03}           & \textbf{99}  & 319          & 459          & 475          & 786 & 120          \\ \hline
\textit{Matyas}           & \textbf{251} & 374          & 366          & 284          & 623 & 344          \\ \hline
\textit{Price02}          & 634          & 522          & \textbf{442} & 461          & 814 & 509          \\ \hline
\textit{Quadratic}        & \textbf{261} & 356          & 378          & 339          & 622 & 331          \\ \hline
\textit{Rastrigin}        & \textbf{355} & 514          & 482          & 487          & 539 & 513          \\ \hline
\textit{Rosenbrock}       & 504          & 386          & 366          & \textbf{345} & 608 & 442          \\ \hline
\textit{Rotatedellipse01} & \textbf{245} & 380          & 355          & 250          & 621 & 372          \\ \hline
\textit{Salomon}          & 510          & 506          & \textbf{458} & 513          & 828 & 606          \\ \hline
\textit{Schaffer03}       & 515          & 586          & \textbf{466} & 475          & 810 & 478          \\ \hline
\textit{Schaffer04}       & 626          & 498          & \textbf{474} & 483          & 826 & 582          \\ \hline
\textit{Schwefel04}       & 422          & 642          & 450          & 467          & 822 & \textbf{304} \\ \hline
\textit{Treccani}         & \textbf{283} & 350          & 389          & 296          & 622 & 362          \\ \hline
\textit{Ursem04}          & \textbf{201} & 482          & 458          & 541          & 814 & 536          \\ \hline
\textit{Whitley}          & 459          & 488          & 482          & 459          & 810 & \textbf{382} \\ \hline
\textit{Zettl}            & \textbf{312} & 398          & 350          & 339          & 623 & 333          \\ \hline
\textit{Zirilli}          & 395          & 410          & \textbf{350} & 369          & 620 & 379          \\ \hline
Lowest count & \textbf{11} & 3 & 6 & 4 & 0 & 5 \\ \hline

\end{tabular}
\label{table:mean_fitness_inv_without_hpo}
\end{table}
Figure~\ref{fig:iterations} illustrates the cHM optimisation process for different benchmark functions (\textit{Ackley02}, \textit{Price02}, and \textit{Rastrigin}) for 50 experiment repetitions. The plots confirm that the optimisation process highly depends on an initialisation. There are cases where the problem is nearly solved after initialisation, and cHM needs only one iteration to find the optimal solution. On the other hand, when the initialisation process is less fortunate, cHM may get stuck in a certain area of the search space (see Figure~\ref{fig:iterations}.

Generally speaking, three types of cHM optimisation patterns occurred for different benchmark functions, all demonstrated on the Figure ~\ref{fig:iterations}. Firstly, there were functions (e.g. \textit{Rastrigin}) for which cHM usually found a near-optimal solution within the first phase of the algorithm run. Secondly, taking the \textit{Ackley02} as an example, for some functions, cHM required approximately two iterations to reach the optimal value for certain functions, iteratively decreasing the cost function. Finally, the third optimisation pattern, illustrated for the \textit{Price02} benchmark, reveals cHM's capability to solve complex problems when the optimisation process uses all algorithm iterations (3 iterations for Figure ~\ref{fig:iterations}). It is crucial to note that for the latter two scenarios, cHM utilised various inner-metaheuristics, indicating its ability to leverage various optimisation techniques within the algorithm. These results demonstrate that cHM can solve simple and complex problems and provides a suitable optimisation tactic for a given task.

\section{Practical application}
\label{sec:Application}

One area in which metaheuristic methods have been extensively studied is feature selection in machine learning
\cite{metas_for_fs_survey_1}, 
\cite{metas_for_fs_survey_2}. 
In this task, exhaustive searching of the entire space is computationally 
prohibitive and leads to exponential growth in search time with increasing numbers of attributes \cite{pso_for_fs_1},
\cite{ga_for_fs_1}. 
Swarm-based algorithms are well-known for their efficient global search capabilities \cite{pso_for_fs_2}, 
which makes 
cHM 
a natural approach to solving feature selection problem.

To show the advantage of the cHM metaheuristic selection mechanism, 
the method was compared with single component metaheuristics (SA, PSO, GA, DE, and BFO) on three well-known classification benchmarks. 
The following parameters of the cHM procedure were used for 
considered datasets:

\begin{itemize}
    \item Loan dataset \cite{loan_kaggle}: $maxFE_{fit}=400$ and $maxFE_{probing}=800$, 
    \item Heart dataset \cite{heart_disease_45}: $maxFE_{fit}=600$ and $maxFE_{probing}=1200$,    
    \item Diabetic Retinopathy dataset \cite{diabetic_retinopathy_debrecen_329}: $maxFE_{fit}=600$ and $maxFE_{probing}=1200$. 
\end{itemize}

For all three benchmarks, $n=4$ was used. 
Each experiment was repeated 10 times for each metaheuristic method. 
A single 
run was constrained by 
the sum of $maxFE_{fit}$ and $maxFE_{probing}$ function evaluations in each optimisation iteration. 
For all methods, the population size was set to $10$ individuals. 
The cost function was calculated as the performance of a classifier trained on a feature subset.

Classification was performed using \textit{RandomForestClassifier} (\textit{RFC}) from the scikit-learn package with 50 inner-estimators (n\_estimators parameter) \cite{scikit-learn}. Next, the feature selection mechanism was performed following \cite{ga_for_fs_orig}, with the limitation of the cost function to be only the error rate value (1 - \textit{accuracy}, calculated as in \cite{scikit-learn}). 
For each cHM algorithm iteration, an $RFC$ was trained on the selected feature subset and evaluated using the error rate function. 
Subsequently, this data was used to calculate the number of cost function evaluations used internally in the cHM algorithm ($maxFE_{fit}$ and $maxFE_{probing}$). Besides these differences, cHM operated similarly to the function optimisation process described in Sections \ref{sec:cHM} and \ref{sec:experimental_setup}.
Finally, the error value was calculated on the test set, being a 30\% of the original data size, and averaged over 10 experiment runs.

The results are presented in Tables ~\ref{table:fs_application_table_loan} - ~\ref{table:fs_application_table_diabetic}.
In general, 
the classification performance can be improved by pruning features 
with the help of the proposed 
cHM method. Indeed, the average error rate of cHM was smaller for all tested benchmarks 
than for all other methods. 
The standard deviation of the error value was the lowest for 
PSO followed by cHM.

On the other hand, the average number of features used for classification was often lower for other methods than for the cHM. 
Although fewer features generally mean lower computation cost, they may cause a performance loss by pruning features with information that is significant for data classification. The primary measure of a method's performance for the studied problem was the error rate. 
Consequently, the cost function did not directly account for the feature-pruning effect. 

\begin{table}[]
\caption{Error rate values with and without feature selection for the \textit{RFC} for the Loan dataset (Kaggle).}
\begin{tabular}{|c|c|c|c|c|c|}
\hline
meta\_name & avg\_cost           & std\_cost            & avg\_num\_features & median\_num\_features & std\_num\_features \\ \hline
BFO        & 0.177 & 0.0145 & 5.2 & \textbf{5}                & 1.932 \\ \hline
DE         & 0.158 & 0.0114 & \textbf{5.0} & \textbf{5}                & 2.119  \\ \hline
GA         & 0.145 & 0.0088 & 6.2 & 7                & 1.075 \\ \hline
cHM        & \textbf{0.143} & 0.0057 & 7.1 & 7                & 1.135 \\ \hline
PSO        & 0.156 & \textbf{0.0047} & 9.1 & 8                & \textbf{0.943} \\ \hline
SA         & 0.185 & 0.0248  & 5.6 & \textbf{5}                & 1.595 \\ \hline
None       & 0.180                & -             & 11    & 11                 & -                \\ \hline
\end{tabular}
\label{table:fs_application_table_loan}
\end{table}

\begin{table}[]
\caption{Error rate values with and without feature selection for the \textit{RFC} for the Heart Disease (Cleveland) dataset (UCI).}
\begin{tabular}{|c|c|c|c|c|c|}
\hline
meta\_name & avg\_cost & std\_cost & avg\_num\_features & median\_num\_features & std\_num\_features \\ \hline
BFO & 0.139 & 0.022 & \textbf{6.7}  & \textbf{7}  & 1.889 \\ \hline
DE  & 0.094 & 0.017 & 7.0  & \textbf{7}  & 1.414 \\ \hline
GA  & 0.093 & 0.016 & 7.6  & 8  & 1.578 \\ \hline
cHM & \textbf{0.089} & 0.016 & 9.1  & 9  & 2.183 \\ \hline
PSO & 0.100   & \textbf{0.013} & 10.4 & 11 & \textbf{1.265} \\ \hline
SA  & 0.144 & 0.026 & 7.2  & \textbf{7}  & 1.135 \\ \hline
None       & 0.144 & -            & 13      & 13                 & -                \\ \hline
\end{tabular}
\label{table:fs_application_table_heart}
\end{table}

\begin{table}[]
\caption{Error rate values with and without feature selection for the \textit{RFC} for the Diabetic Retinopathy dataset (UCI).}
\begin{tabular}{|c|c|c|c|c|c|}
\hline
meta\_name & avg\_cost & std\_cost & avg\_num\_features & median\_num\_features & std\_num\_features \\ \hline
BFO & 0.292 & 0.010  & 8.6  & \textbf{9}  & 1.776 \\ \hline
DE  & 0.268 & 0.010  & 8.8  & 10 & 2.616 \\ \hline
GA  & 0.275 & 0.008 & 8.7  & \textbf{9}  & 1.829 \\ \hline
cHM & \textbf{0.267} & \textbf{0.004} & 8.7  & \textbf{9}  & 2.541 \\ \hline
PSO & 0.285 & \textbf{0.004} & 15.7 & 16 & \textbf{1.494} \\ \hline
SA  & 0.296 & 0.006 & \textbf{8.0}  & \textbf{9}  & 2.055 \\ \hline
None       & 0.321     & -   & 19    & 19                 & -                \\ \hline
\end{tabular}
\label{table:fs_application_table_diabetic}
\end{table}

%
%
\section{Conclusions}
\label{sec:Conclusions}

The proposed cHM algorithm demonstrates superior performance compared to component metaheuristics. Although performance depends on a particular benchmark, as shown in Tables~\ref{table:mean_fitness_without_hpo} -~\ref{table:min_fitness_without_hpo}, the aggregated performance is significantly higher for the cHM algorithm, as highlighted in Table~\ref{table:aggregated_euclidean_fitness_distance}.
Table~\ref{table:iteration_chm_single_merged}  shows that cHM successfully integrates weak and strong population-based techniques to form an effective hybrid strategy. Although the cHM most frequently selected the DE algorithm, the proportion of iterations employing other methods remained substantial. Furthermore, these tables and  Figure~\ref{fig:iterations} confirm that cHM enables efficient transitions of populations across algorithmic phases and iterations. In addition, the figure illustrates that component metaheuristics are shuffled and selected based on the specific requirements of each optimisation phase, enhancing the method’s flexibility and effectiveness. As described in Tables~\ref{table:mean_fitness_without_hpo} -~\ref{table:min_fitness_without_hpo}, the performance of population-based techniques heavily depends on a particular benchmark problem. 
Conversely, cHM offers a universal solution approach for all tested optimisation tasks. The method is particularly useful in scenarios where metaheuristic selection is challenging or resource-intensive, as it uses the same or lower number of fitness invocations than other methods to achieve the same or higher performance (Table~\ref{table:mean_fitness_inv_without_hpo}). On the downside, the method exhibits sensitivity to initialisation and generally has a high standard deviation across experiment repetitions (Table~\ref{table:std_fitness_without_hpo}). 
The cHM algorithm was successfully applied to the feature selection problem for classification data. The results presented in Tables \ref{table:fs_application_table_loan} - \ref{table:fs_application_table_diabetic} demonstrate that the cHM significantly decreased the error rate after applying feature selection algorithms. In addition, across the three considered datasets, the proposed hybrid method outperformed individual component metaheuristics proving its practical usefulness and efficiency. 

\subsubsection*{Future work}
The cHM algorithm can be extended with adaptive migration strategies driven by machine learning, where reinforcement learning or bandit-based controllers dynamically regulate the allocation of inner metaheuristics during different phases of optimisation. Incorporating problem-specific features into the selection process would allow cHM to adapt not only to population dynamics but also to structural characteristics of the optimisation landscape.

Scalability of cHM remains a central direction. Evaluation in large-scale distributed environments, including cloud and GPU-based infrastructures, will make it possible to assess its efficiency in practical applications. Hybridisation with surrogate models offers another opportunity, reducing evaluation costs in computationally expensive black-box problems and widening the scope of applicability of cHM.

The framework also requires a deeper analysis of diversity metrics as part of its selection and switching mechanisms. Beyond standard population-based indicators, the use of information-theoretic and landscape-aware measures could refine the estimation of exploration and exploitation balance. Integrating such metrics with multi-objective strategies inside cHM would strengthen its generality and robustness, creating a more adaptive universal optimiser.

\section*{Acknowledgements}
\noindent
Jacek Ma{\'n}dziuk was partially supported by the National Science Centre, Poland, grant number 2023/49/B/ST6/01404. 
The research project was partially supported by the program „Excellence Initiative – research university” for the AGH University of Krakow and was partially supported by a Grant for Statutory Activity from the Faculty of Physics and Applied Computer Science of the AGH.





\clearpage 





\newpage
 \section*{Appendix}

\appendix

\begin{table}[H!]
\centering
\caption{Definitions of the benchmark functions.}
\begin{tabular}{|c|c|c|}
\hline
\textit{Function name}      & Equation                                                                                         & Minimum $(x_1^*, x_2^*)$ \\ \hline
\textit{Ackley02}           & $-200\exp(-0.02\sqrt{x_1^2 + x_2^2})$                                                            & $(0, 0)$                 \\ \hline
\textit{Beale}              & $(x_1 x_2 - x_1 + 1.5)^2 + (x_1 x_2^2 - x_1 + 2.25)^2 + (x_1 x_2^3 - x_1 + 2.625)^2$             & $(3, 0.5)$               \\ \hline
\textit{Bird}               & $(x_1 - x_2)^2 + \exp((1-\sin(x_1))^2)\cos(x_2) + \exp((1-\cos(x_2))^2)\sin(x_1)$                & $(4.70106, 3.15295)$     \\ \hline
\textit{Bohachevsky01}      & $x_1^2 + 2x_2^2 - 0.3\cos(3\pi x_1) - 0.4\cos(4\pi x_2) + 0.7$                                   & $(0, 0)$                 \\ \hline
\textit{Branin02} & 
$(x_2 - \frac{5.1}{4\pi^2}x_1^2 + \frac{5}{\pi}x_1 - 6)^2 + 10(1 - \frac{1}{8\pi})\cos(x_1) + 10$ & 
(-3.2, 12.53) \\ \hline
\textit{Brent}              & $(x_1 + 10)^2 + (x_2 + 10)^2 + \exp(-x_1^2 - x_2^2)$                                             & $(-10, -10)$             \\ \hline
\textit{Brown}              & $(x_1^2)^{x_2^2 + 1} + (x_2^2)^{x_1^2 + 1}$                                                      & $(0, 0)$                 \\ \hline
\textit{Eggcrate}           & $x_1^2 + x_2^2 + 25(\sin^2(x_1) + \sin^2(x_2))$                                                  & $(0, 0)$                 \\ \hline
\textit{Goldstein-Price} &
  \begin{tabular}[c]{@{}c@{}}$[1 + (x_1 + x_2 + 1)^2 (19 - 14x_1 + 3x_1^2 - 14x_2 + 6x_1 x_2 + 3x_2^2)] * $\\[1em]$ [30 + (2x_1 - 3x_2)^2 (18 - 32x_1 + 12x_1^2 + 48x_2 - 36x_1 x_2 + 27x_2^2)]$\end{tabular} &
  $(0, -1)$ \\ \hline
\textit{Himmelblau}         & $(x_1^2 + x_2 - 11)^2 + (x_1 + x_2^2 - 7)^2$                                                     & $(3, 2)$                 \\ \hline
\textit{Hosaki}             & $(1 - 8x_1 + 7x_1^2 - \frac{7}{3}x_1^3 + \frac{1}{4}x_1^4)x_2^2 \exp(-x_1)$                      & $(4, 2)$                 \\ \hline
\textit{Keane}              & $\frac{\sin^2(x_1 - x_2)\sin^2(x_1 + x_2)}{\sqrt{x_1^2 + x_2^2}}$                                & $(7.85396153, 7.85396153)$             \\ \hline
\textit{Levy03}             & $\sin^2(\pi y_1) + (y_1 - 1)^2 (1 + 10\sin^2(\pi y_2)) + (y_1 - 1)^2$                            & $(1, 1)$                 \\ \hline
\textit{Matyas}             & $0.26(x_1^2 + x_2^2) - 0.48 x_1 x_2$                                                             & $(0, 0)$                 \\ \hline
\textit{Price02}            & $1 + \sin^2(x_1) + \sin^2(x_2) - 0.1 \exp(-x_1^2 - x_2^2)$                                       & $(0, 0)$                 \\ \hline
\textit{Quadratic}          & $-3803.84 - 138.08x_1 - 232.92x_2 + 128.08x_1^2 + 203.64x_2^2 + 182.25x_1x_2$                    & $(0.19388, 0.48513)$     \\ \hline
\textit{Rastrigin}          & $10[(x_1^2 - 10\cos(2\pi x_2))+(x_2^2 - 10\cos(2\pi x_2))]$                                                                   & $(0, 0)$                 \\ \hline
\textit{Rosenbrock}         & $100(x_1^2 - x_2)^2 + (x_1 - 1)^2$                                                               & $(1, 1)$                 \\ \hline
\textit{Rotated Ellipse 01} & $7x_1^2 - 6\sqrt{3}x_1 x_2 + 13x_2^2$                                                            & $(0, 0)$                 \\ \hline
\textit{Salomon}            & $1 - \cos(2\pi \sqrt{x_1^2+x_2^2}) + 0.1\sqrt{x_2^2+x_2^2}$                                                  & $(0, 0)$                 \\ \hline
\textit{Schaffer03}         & $0.5 + \frac{\sin^2(\cos(|x_1^2 - x_2^2|)) - 0.5}{1 + 0.001(x_1^2 + x_2^2)^2}$                   & $(0, 1.253115)$          \\ \hline
\textit{Schaffer04}         & $0.5 + \frac{\cos^2(\sin(x_1^2 - x_2^2)) - 0.25}{1 + 0.001(x_1^2 + x_2^2)^2}$                    & $(0, 1.253115)$          \\ \hline
\textit{Schwefel04}         & $(x_1 - 1)^2 + (x_2 - x_1^2)^2$                                                                  & $(1, 1)$                 \\ \hline
\textit{Treccani}           & $x_1^4 + 4x_1^3 + 4x_1^2 + x_2^2$                                                                & $(0, 0)$                 \\ \hline
\textit{Ursem04}            & $-3 \sin(0.5\pi x_1 + 0.5\pi) \cdot \frac{2 - \sqrt{x_1^2 + x_2^2}}{4}$                          & $(0, 0)$                 \\ \hline
\textit{Whitley}            & $\frac{(100(x_1^2 - x_2)^2 + (1 - x_2)^2)^2}{4000} - \cos(100(x_1^2 - x_2)^2 + (1 - x_2)^2) + 1$ & $(1, 1)$                 \\ \hline
\textit{Zettl}              & $0.25 x_1 + (x_1^2 - 2x_1 + x_2^2)^2$                                                            & $(-0.02990, 0)$          \\ \hline
\textit{Zirilli}            & $0.25 x_1^4 - 0.5 x_1^2 + 0.1 x_1 + 0.5 x_2^2$                                                   & $(-1.0465, 0)$           \\ \hline
\end{tabular}
\label{table:function_definitions}
\end{table}

\begin{table}[htbp]
\centering
\caption{Grouping of benchmark optimisation functions by landscape characteristics and recommended cHM parameter settings.}
\renewcommand{\arraystretch}{1.0}
\begin{tabular}{|l|c|c|}
\hline
\textbf{Bucket} & 
\textbf{Functions} & 
\begin{tabular}[c]{@{}c@{}}
\textbf{$MaxFE_{probing}$,} \\
\textbf{$MaxFE_{fit}$}
\end{tabular} \\ \hline

\begin{tabular}[t]{@{}l@{}}
\textit{Single-basin}\\[1em]
\textit{(convex or near-convex)}
\end{tabular}
&
\begin{tabular}[t]{@{}l@{}}
Quadratic, Rotatedellipse01,\\[1em]
Matyas, Zirilli
\end{tabular}
& (300, 600) \\ \hline

\begin{tabular}[t]{@{}l@{}}
\textit{Ill-conditioned}\\[1em]
\textit{/ narrow valley}
\end{tabular}
&
\begin{tabular}[t]{@{}l@{}}
Rosenbrock, Brown
\end{tabular}
& (300, 600) \\ \hline

\begin{tabular}[t]{@{}l@{}}
\textit{Highly multimodal}\\[1em]
\textit{(periodic / rippled landscapes)}
\end{tabular}
&
\begin{tabular}[t]{@{}l@{}}
Rastrigin, Ackley02, Salomon, Bohachevsky01,\\[1em]
Eggcrate, Schaffer03, Schaffer04, Levy03,\\[1em]
Schwefel04, Keane, Price02, Ursem04, Whitley
\end{tabular}
& (400, 800) \\ \hline

\begin{tabular}[t]{@{}l@{}}
\textit{Few well-separated}\\[1em]
\textit{global minima (multi-basin)}
\end{tabular}
&
\begin{tabular}[t]{@{}l@{}}
Himmelblau, Branin02, Bird, Treccani,\\[1em]
Goldstein-Price, Beale, Hosaki, Zettl
\end{tabular}
& (300, 600) \\ \hline

\begin{tabular}[t]{@{}l@{}}
\textit{Boundary or shifted}\\[1em]
\textit{minima (simple bowls)}
\end{tabular}
&
\begin{tabular}[t]{@{}l@{}}
Brent
\end{tabular}
& (400, 800) \\ \hline

\end{tabular}
\label{tab:function_buckets_parameters}
\end{table}

\begin{landscape}
\begin{figure}[p]
\centering
\captionsetup{font=small}
\setlength{\tabcolsep}{1.5pt}%
\renewcommand{\arraystretch}{0.96}%

\begin{tabular}{*{7}{c}}
\includegraphics[width=0.18\textwidth]{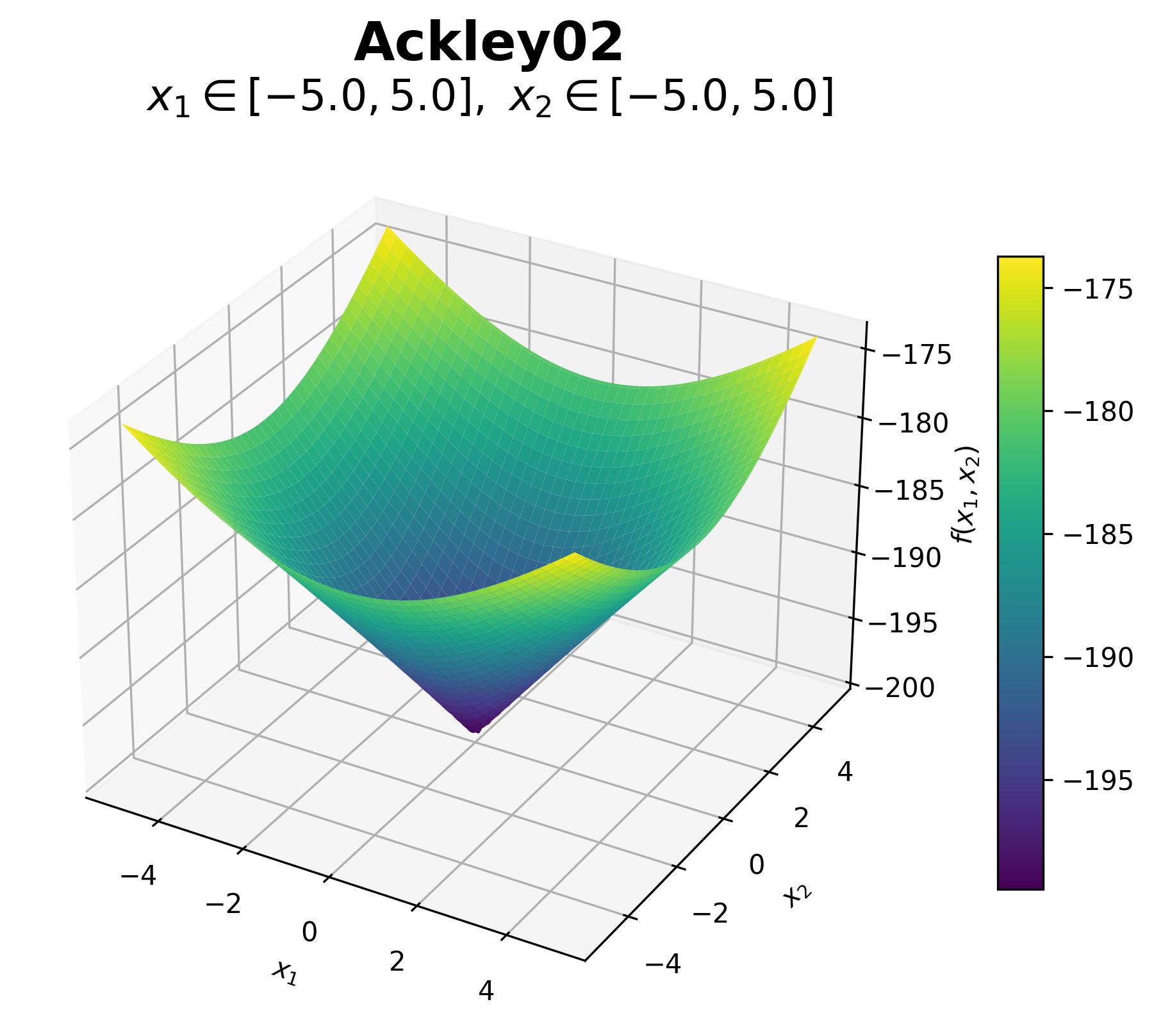} &
\includegraphics[width=0.18\textwidth]{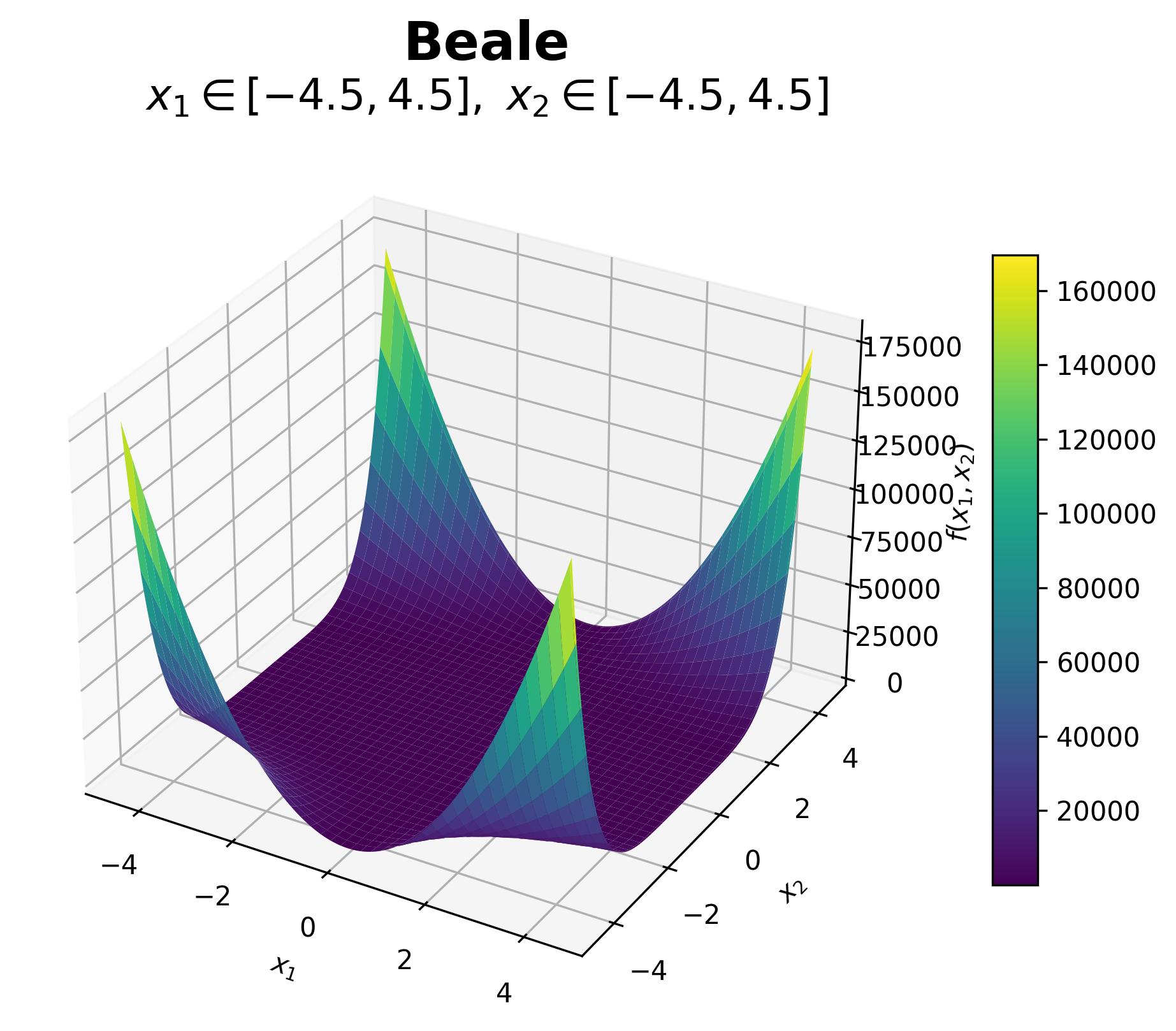} &
\includegraphics[width=0.18\textwidth]{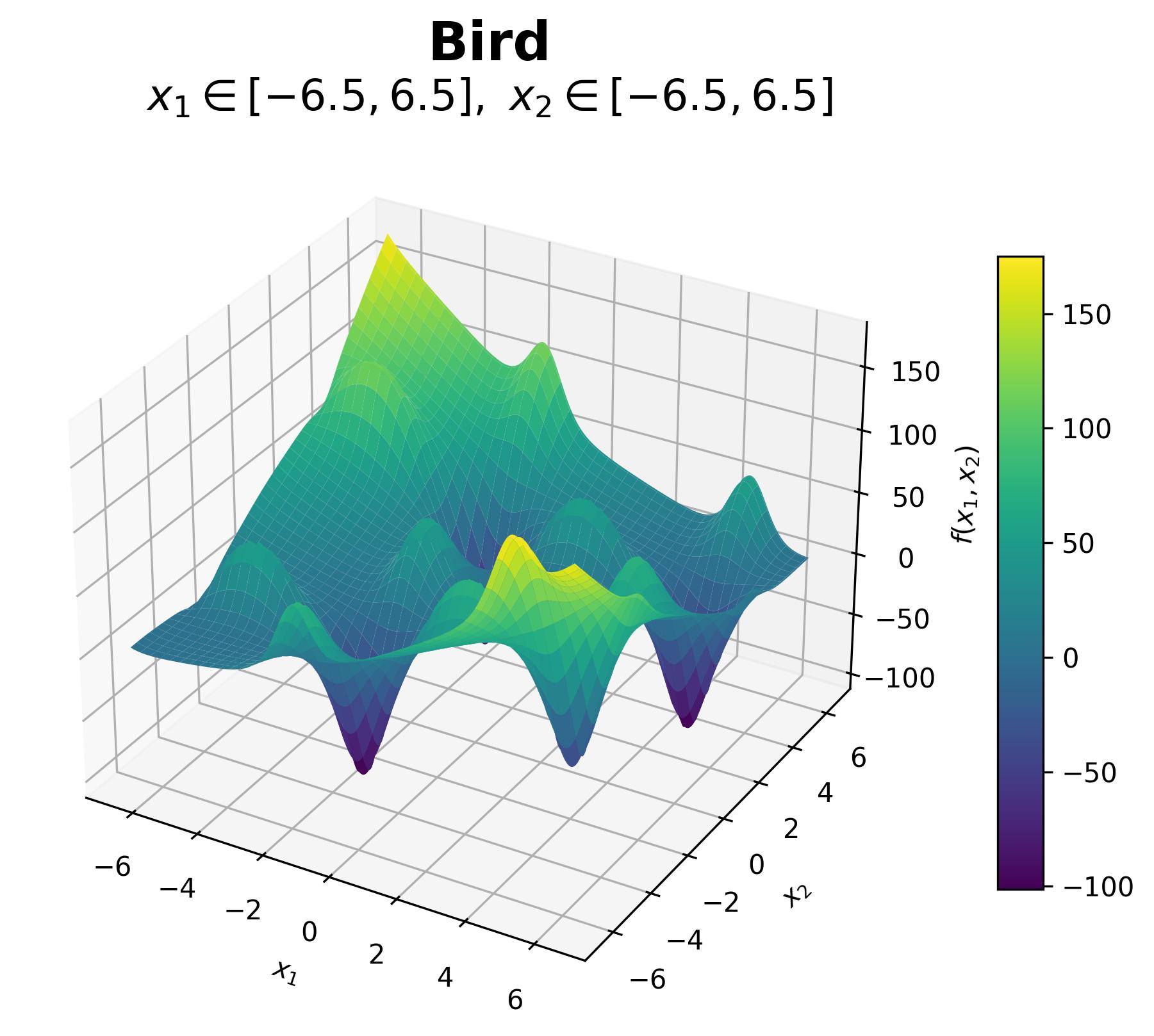} &
\includegraphics[width=0.18\textwidth]{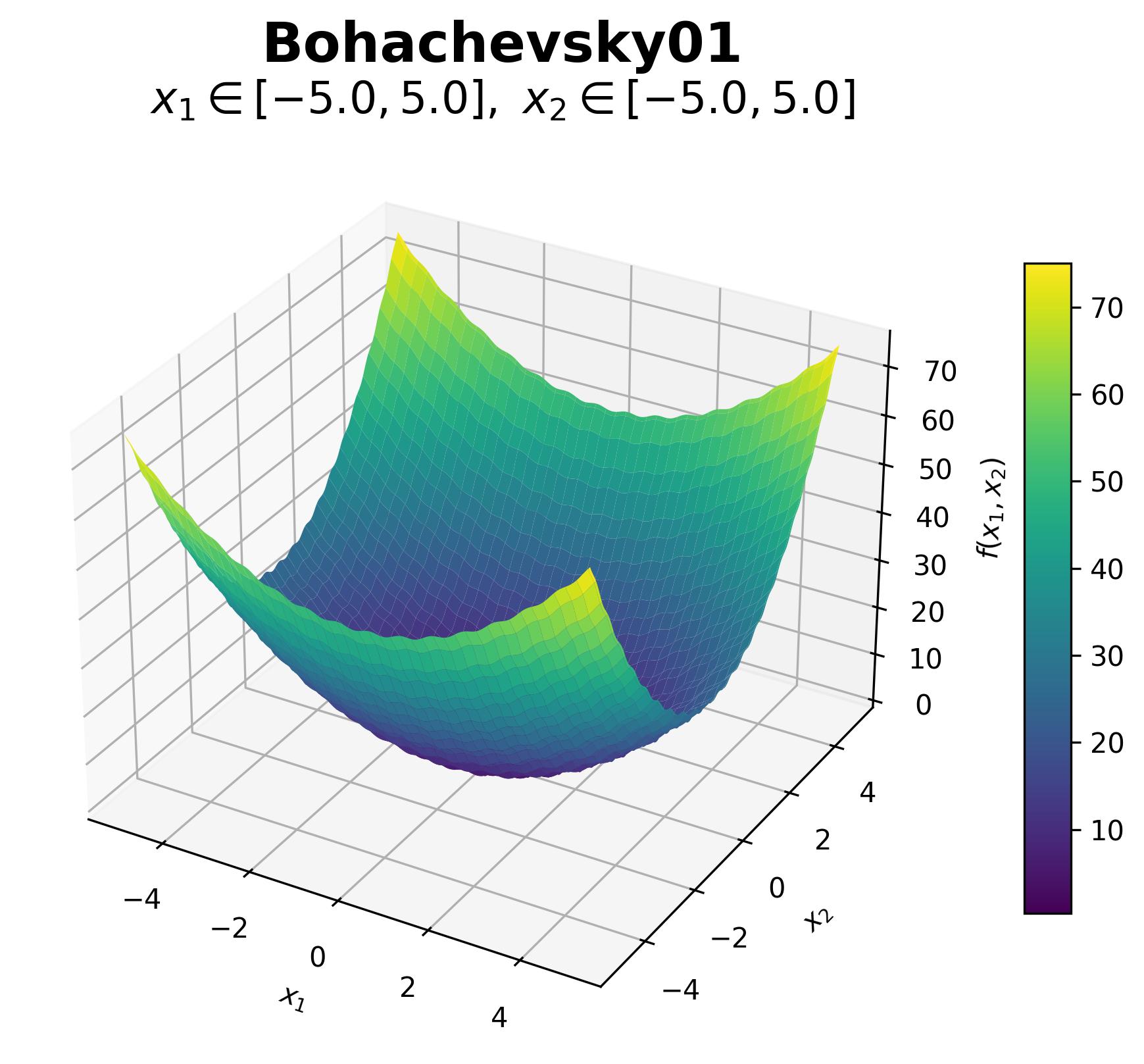} &
\includegraphics[width=0.18\textwidth]{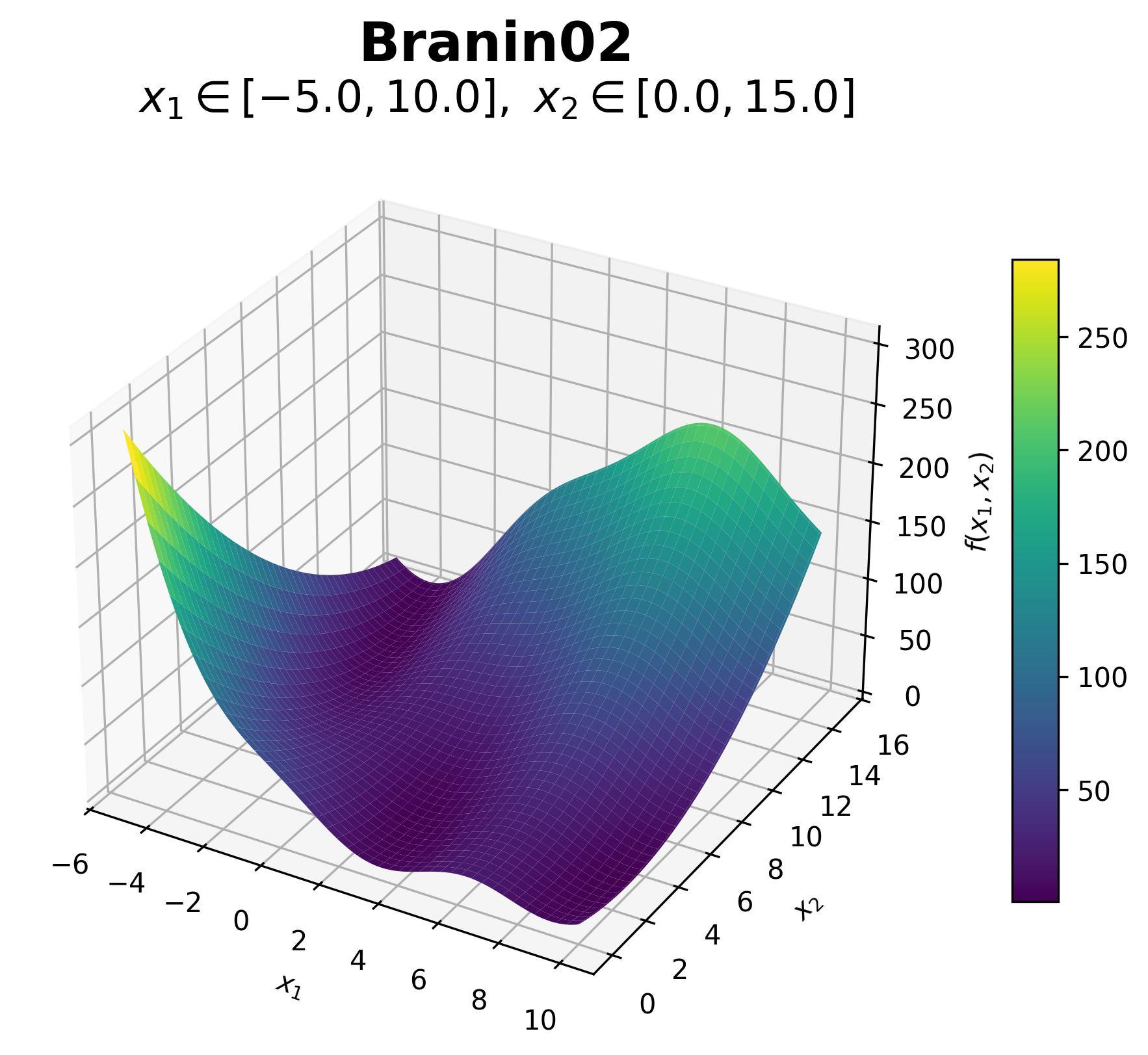} &
\includegraphics[width=0.18\textwidth]{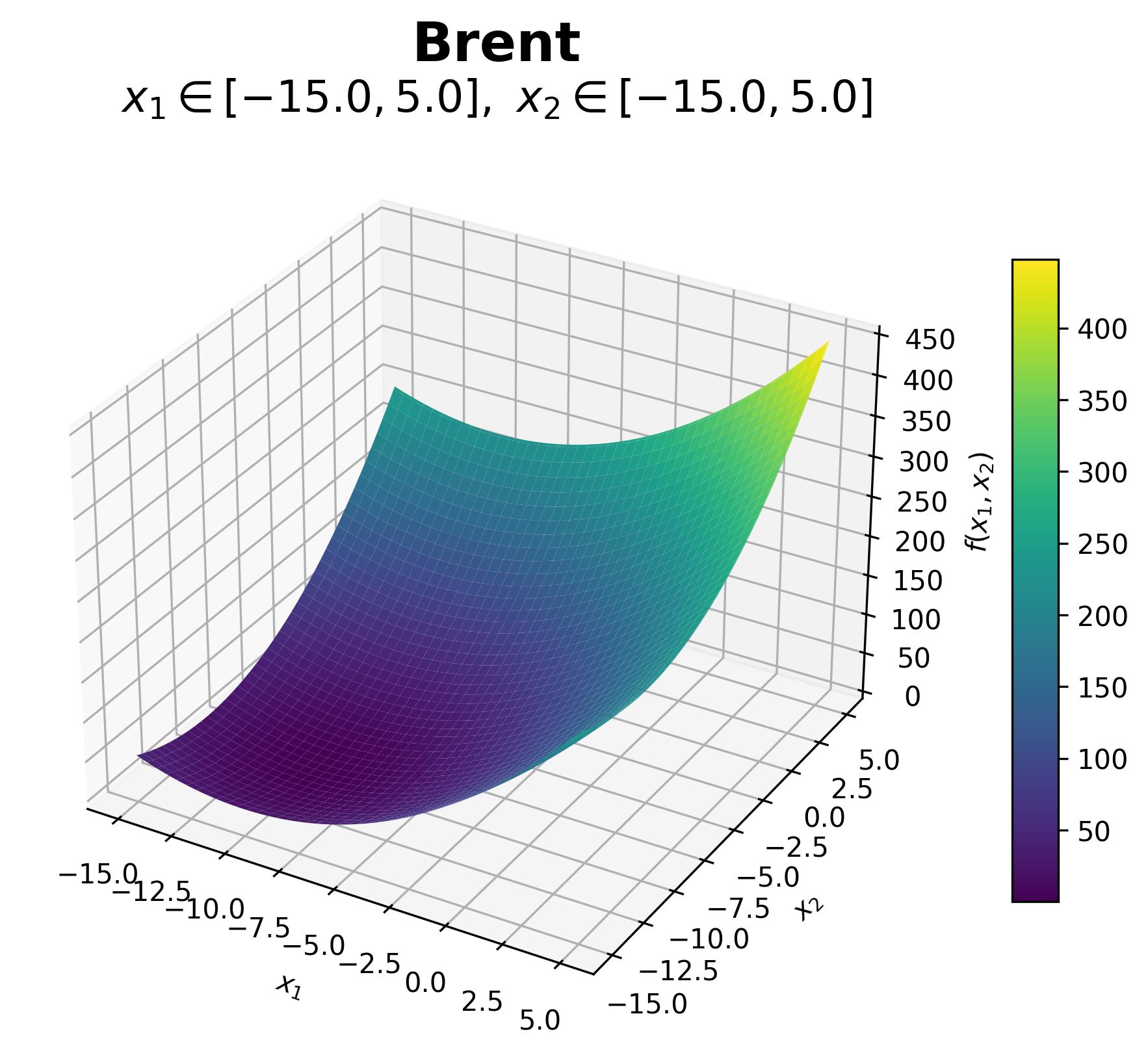} &
\includegraphics[width=0.18\textwidth]{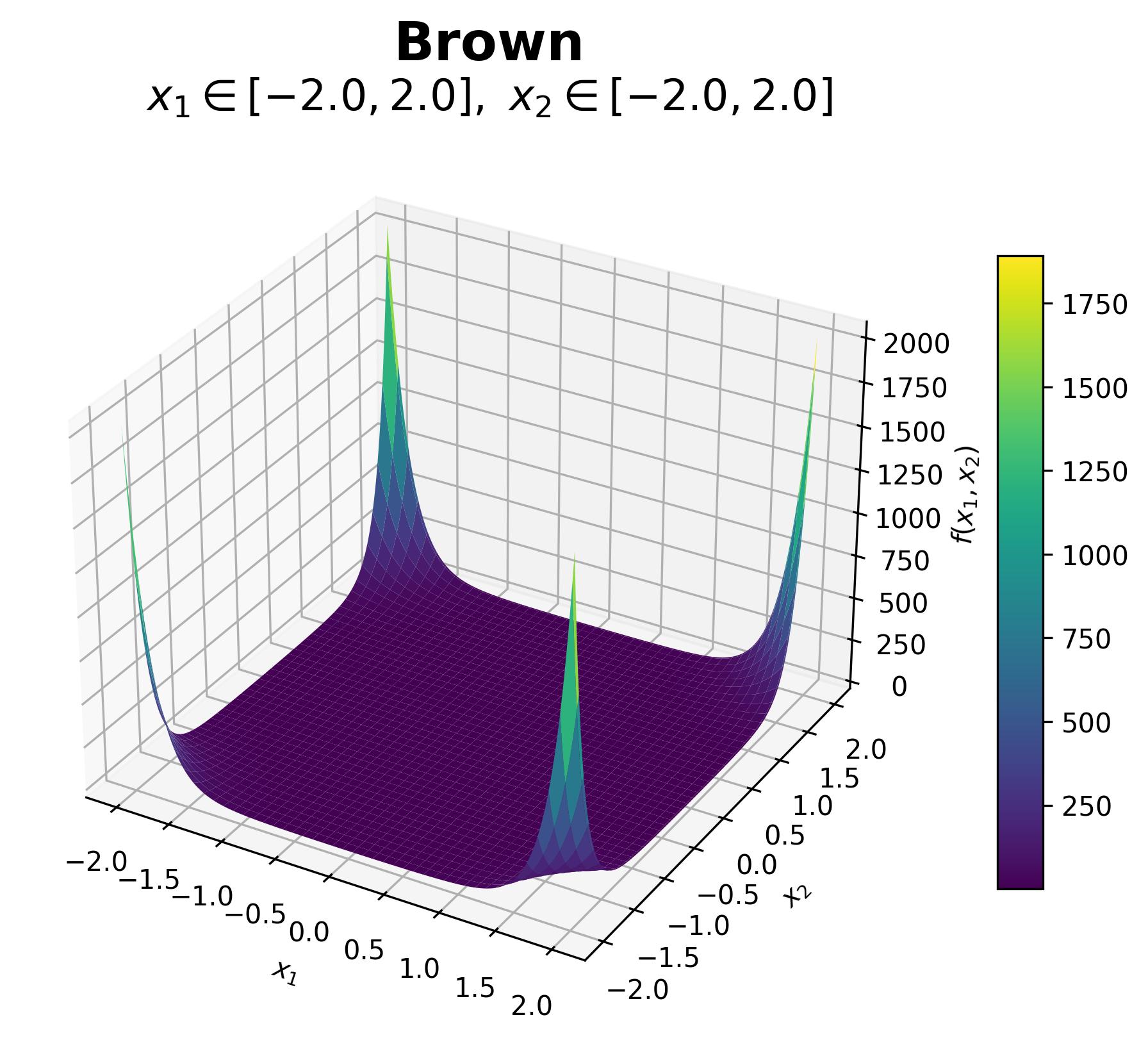} \\
\includegraphics[width=0.18\textwidth]{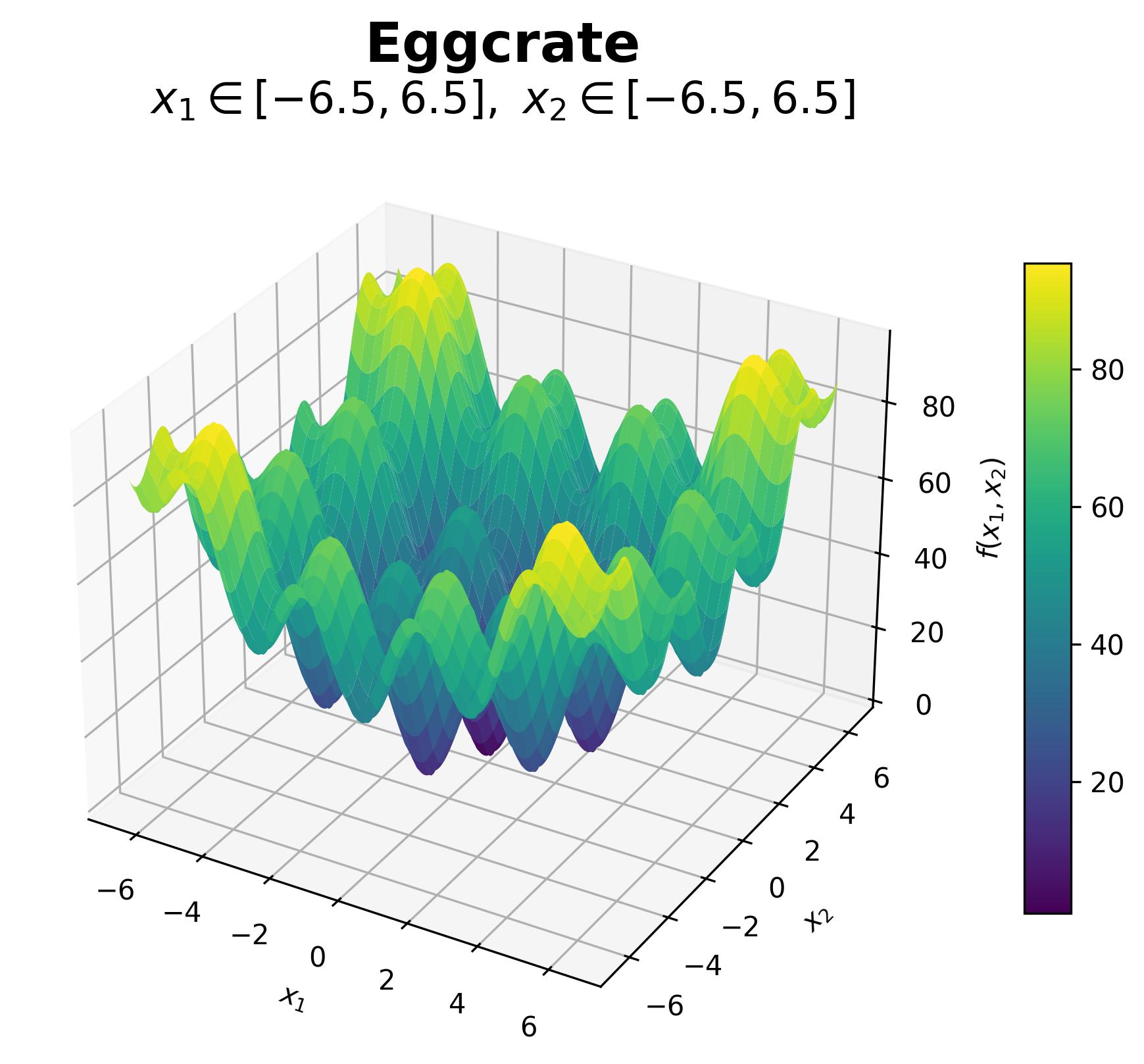} &
\includegraphics[width=0.18\textwidth]{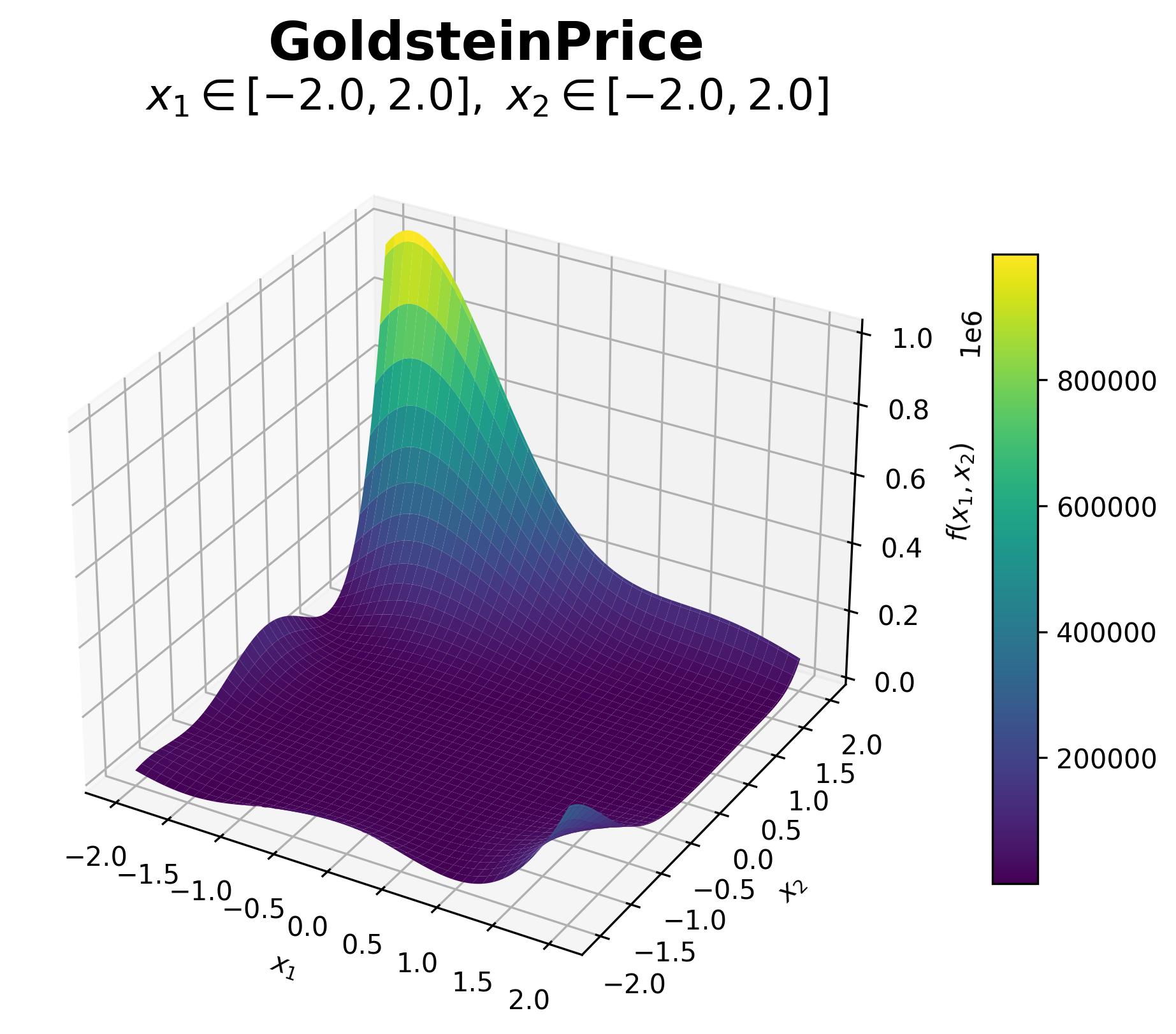} &
\includegraphics[width=0.18\textwidth]{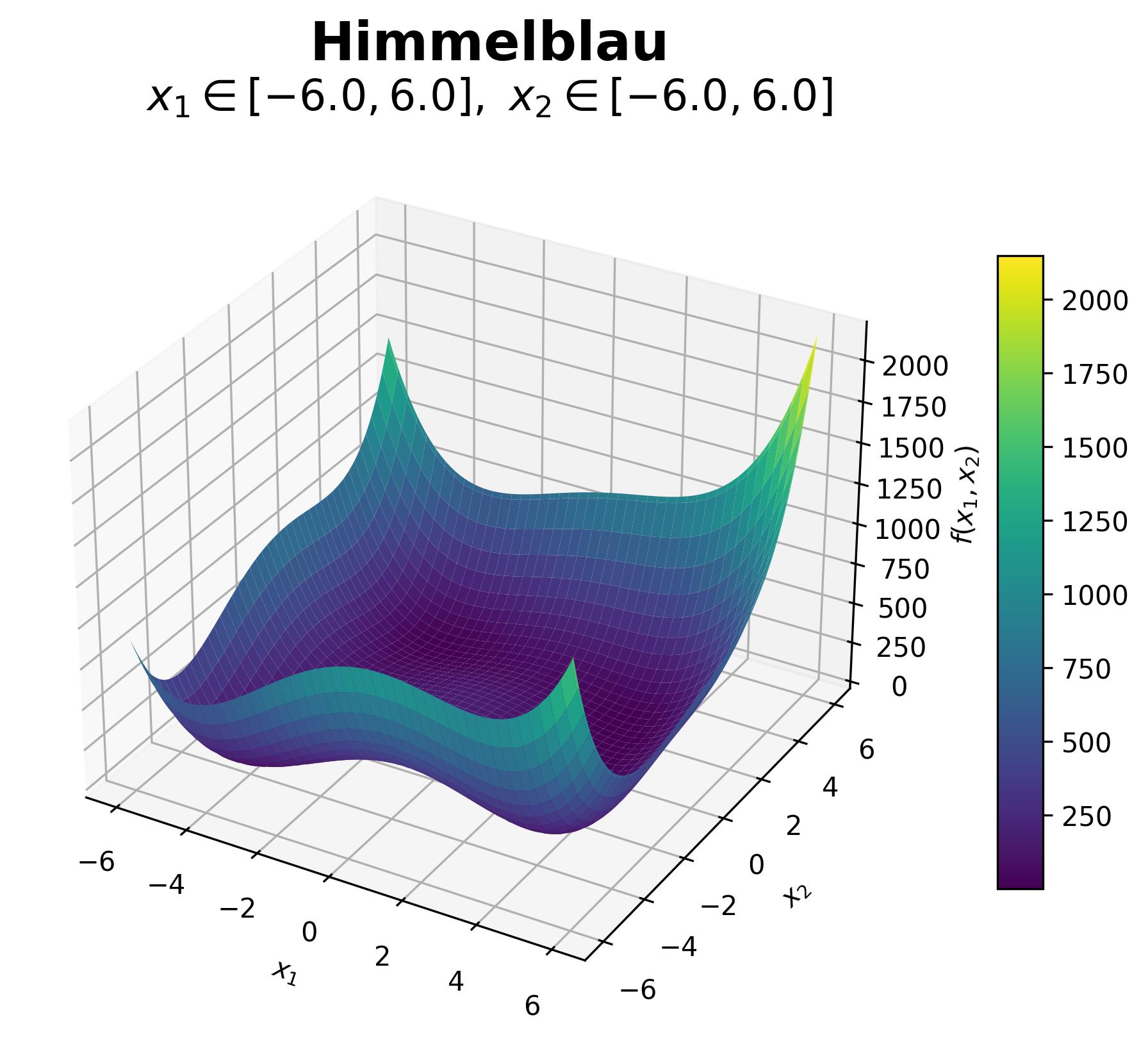} &
\includegraphics[width=0.18\textwidth]{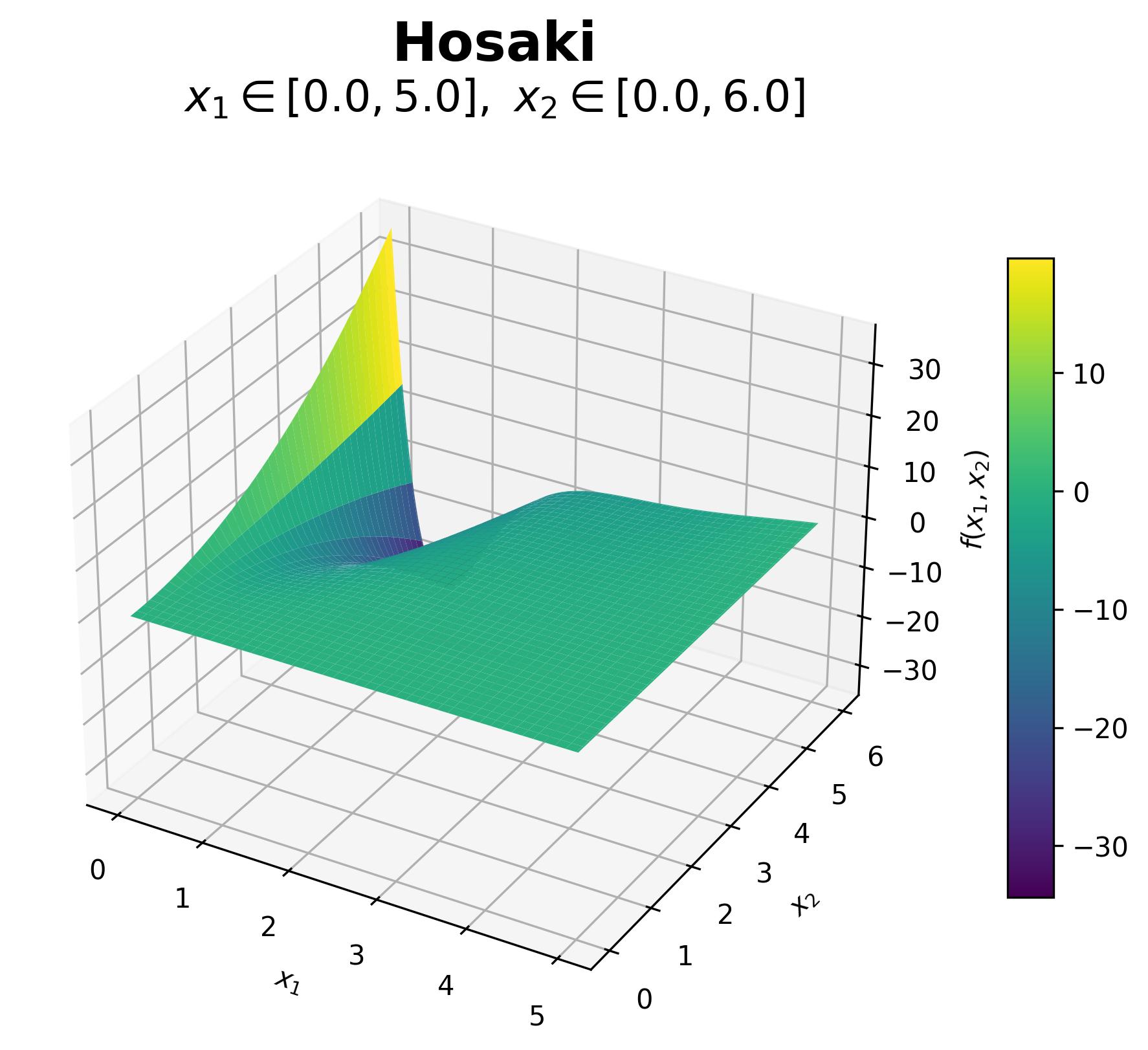} &
\includegraphics[width=0.18\textwidth]{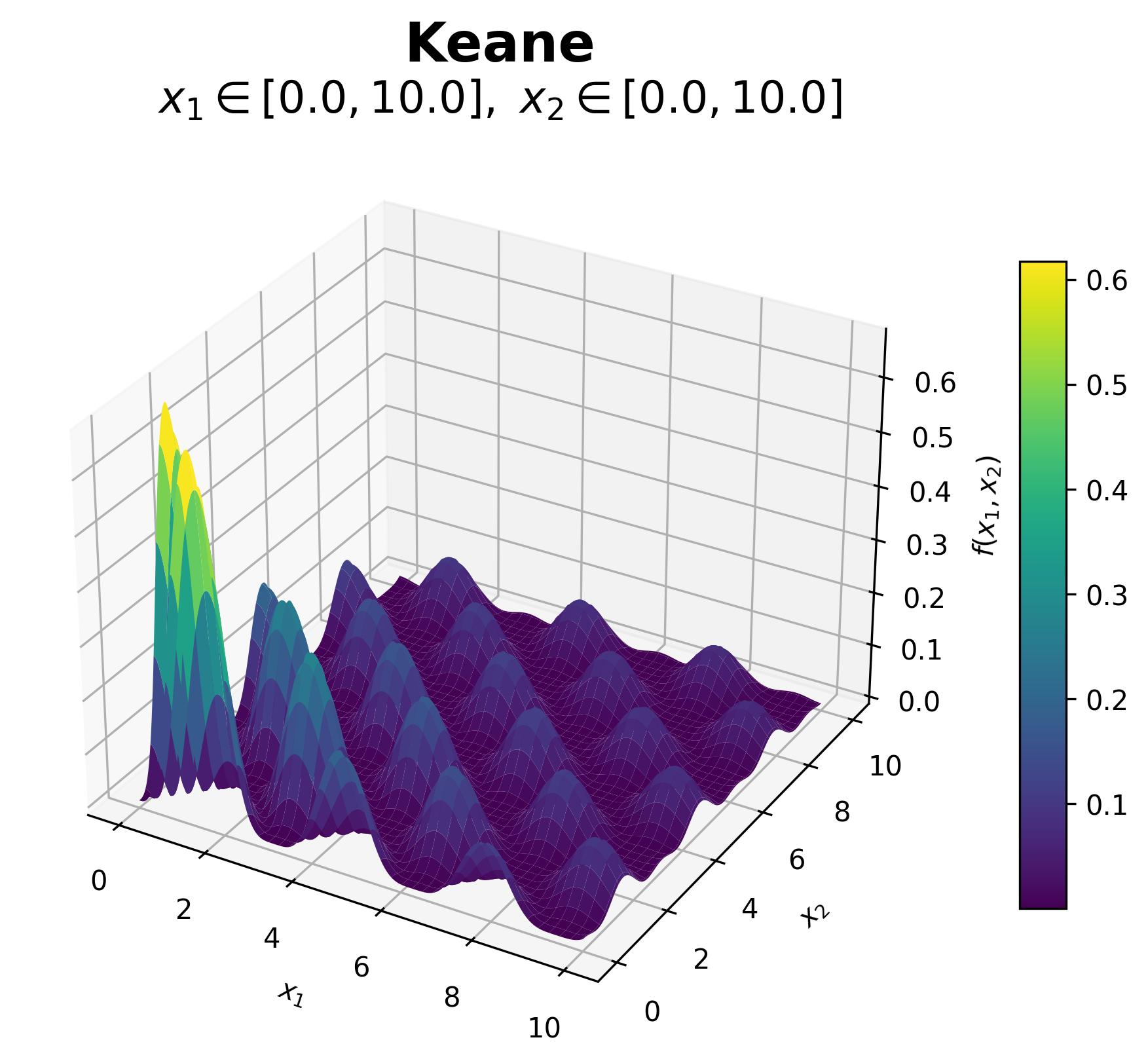} &
\includegraphics[width=0.18\textwidth]{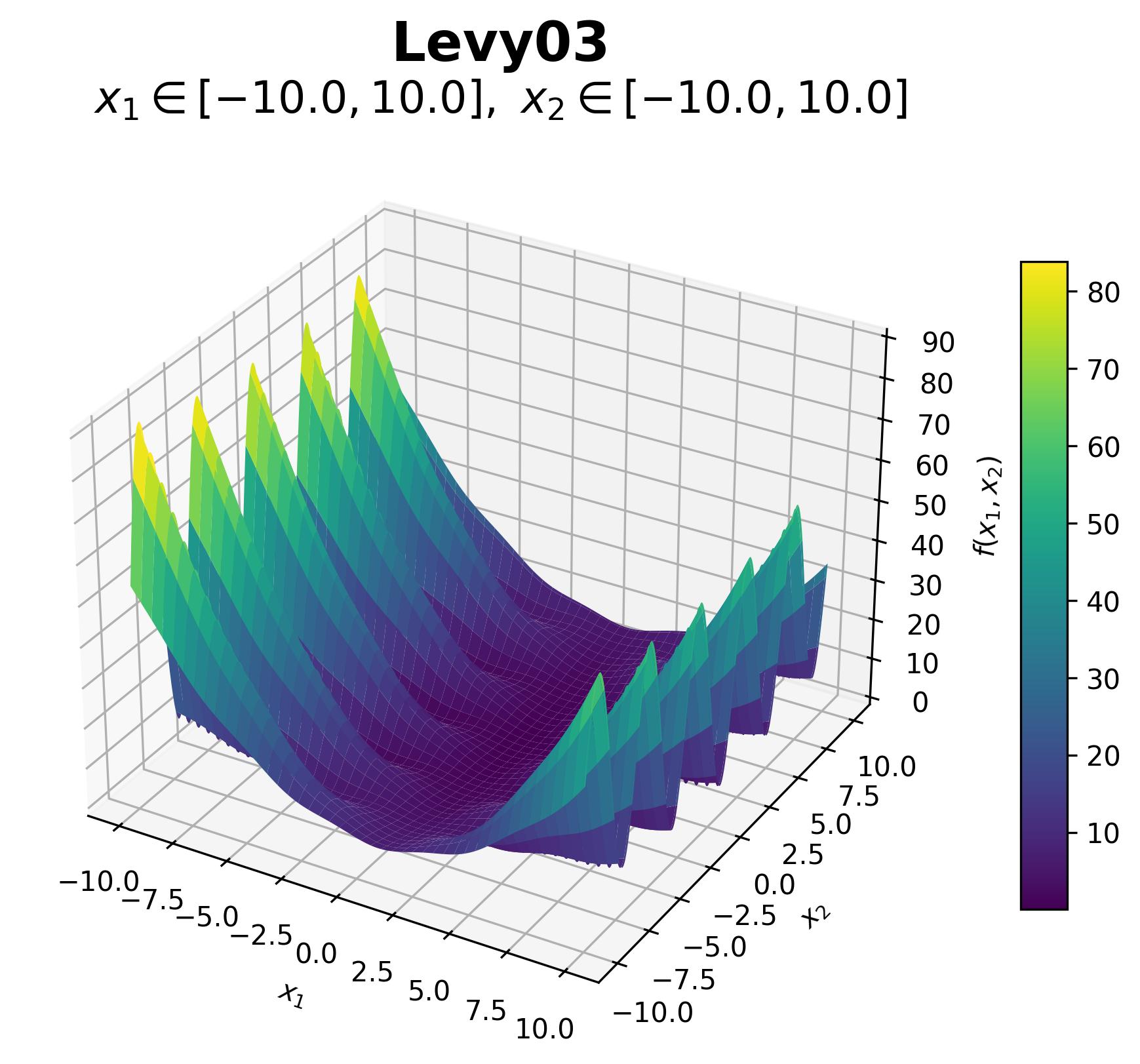} &
\includegraphics[width=0.18\textwidth]{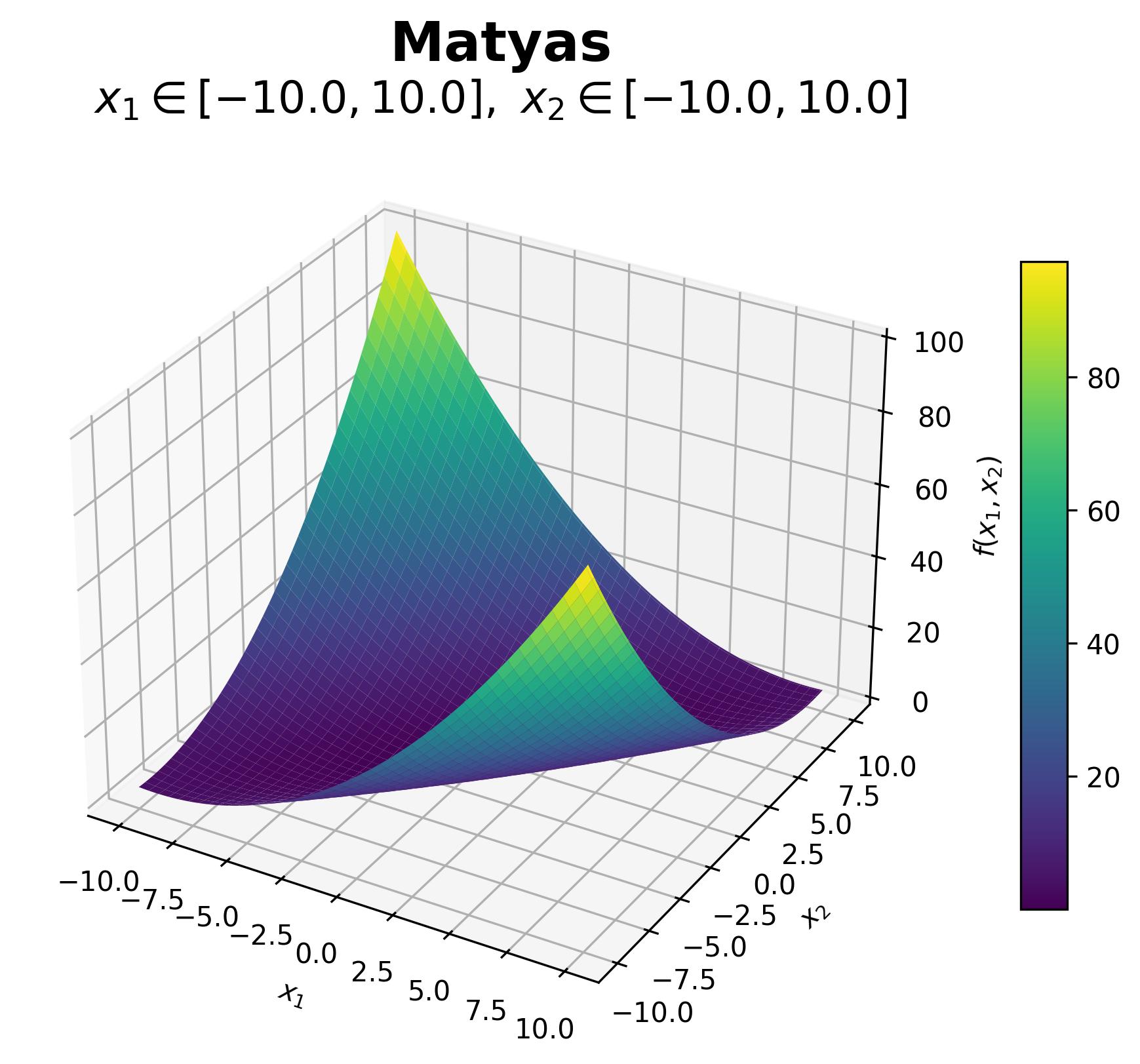} \\
\includegraphics[width=0.18\textwidth]{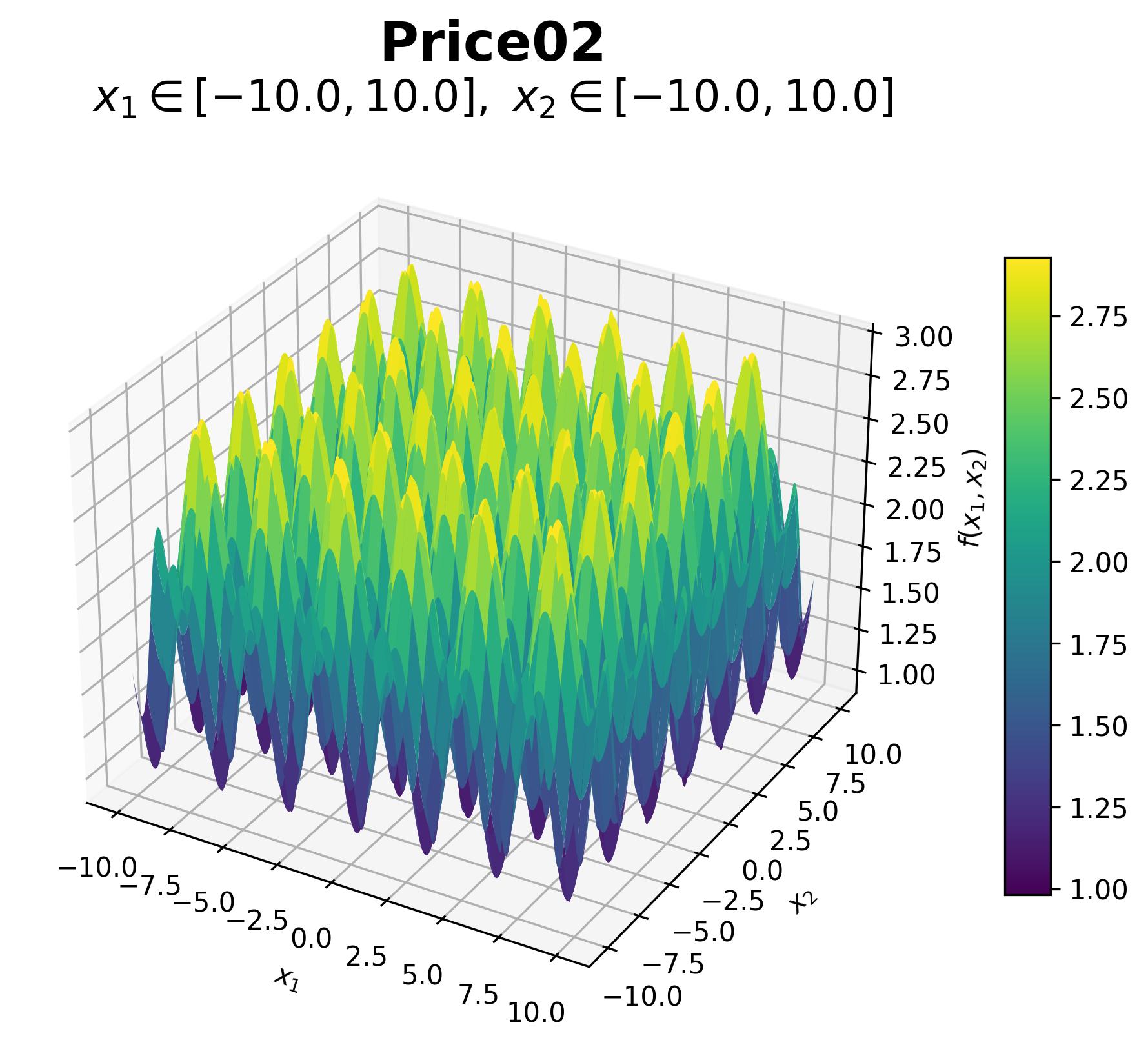} &
\includegraphics[width=0.18\textwidth]{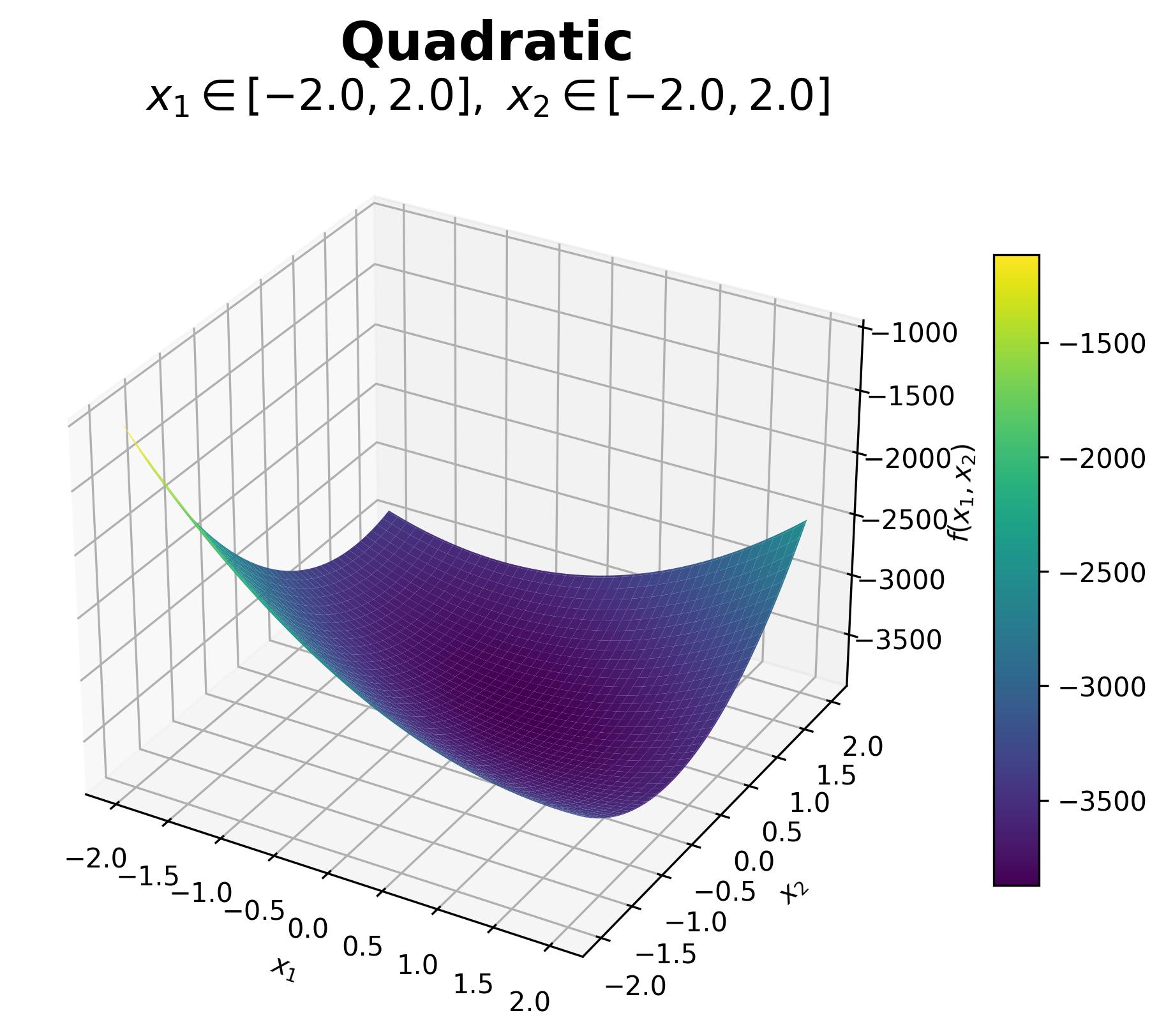} &
\includegraphics[width=0.18\textwidth]{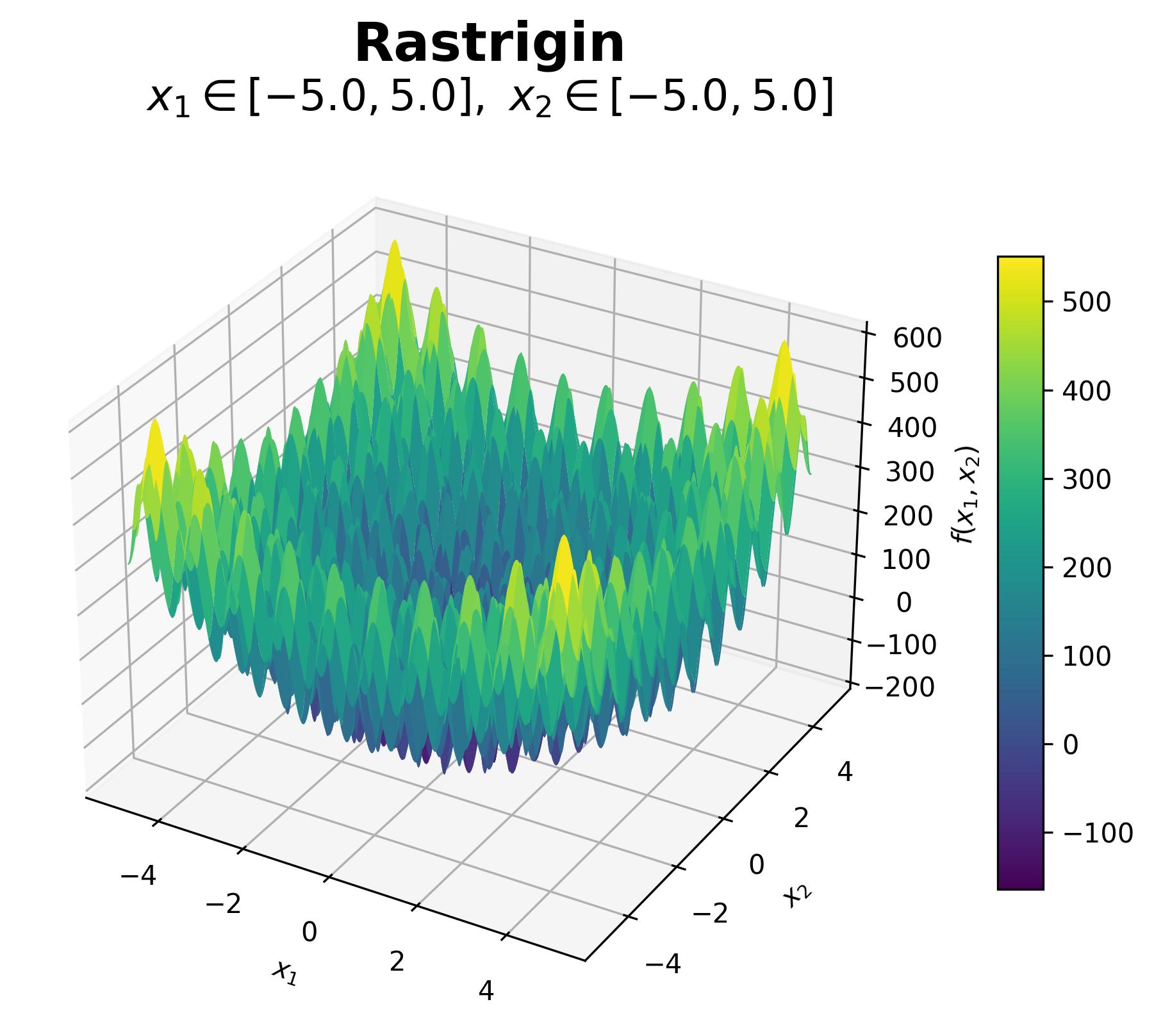} &
\includegraphics[width=0.18\textwidth]{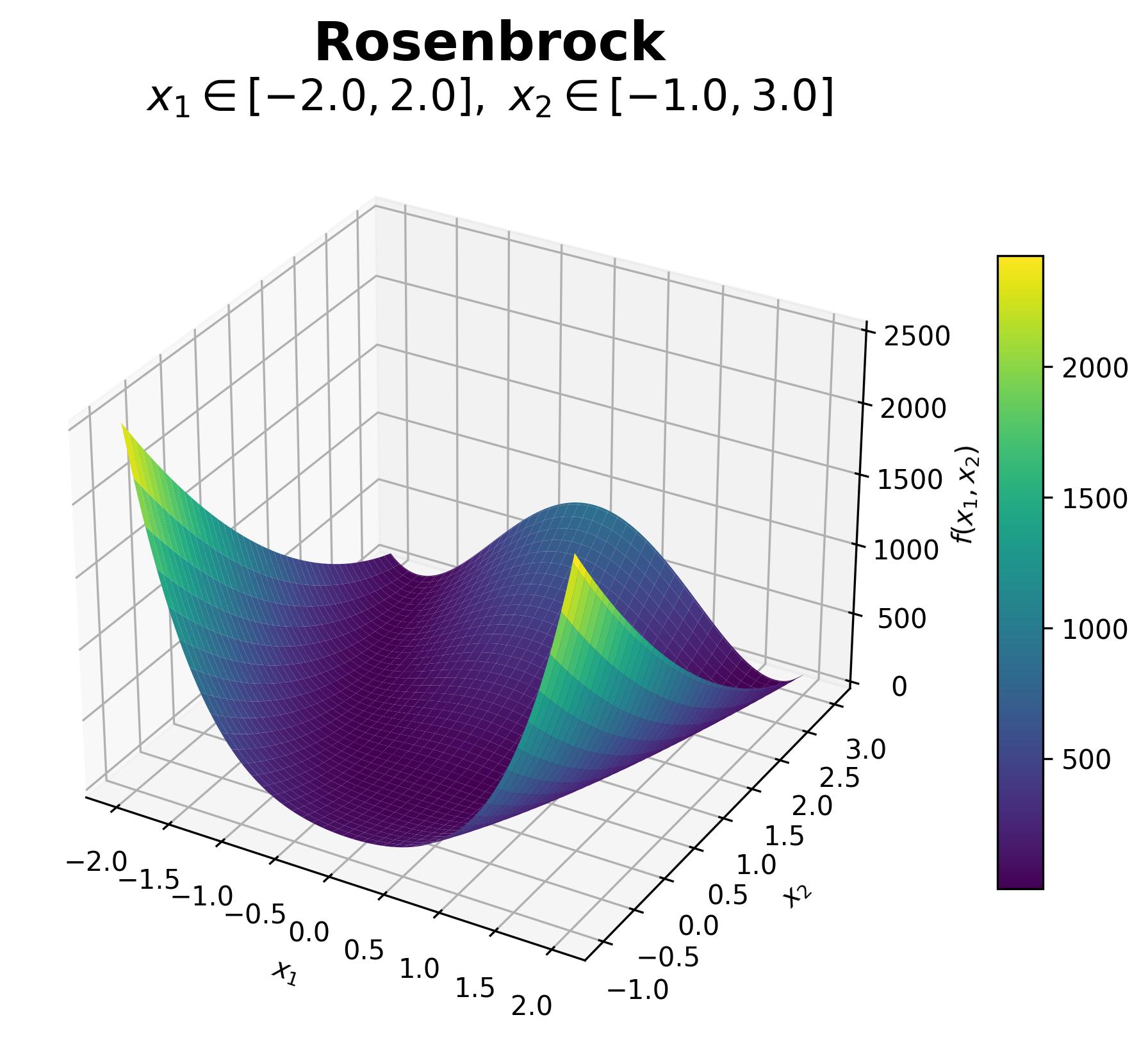} &
\includegraphics[width=0.18\textwidth]{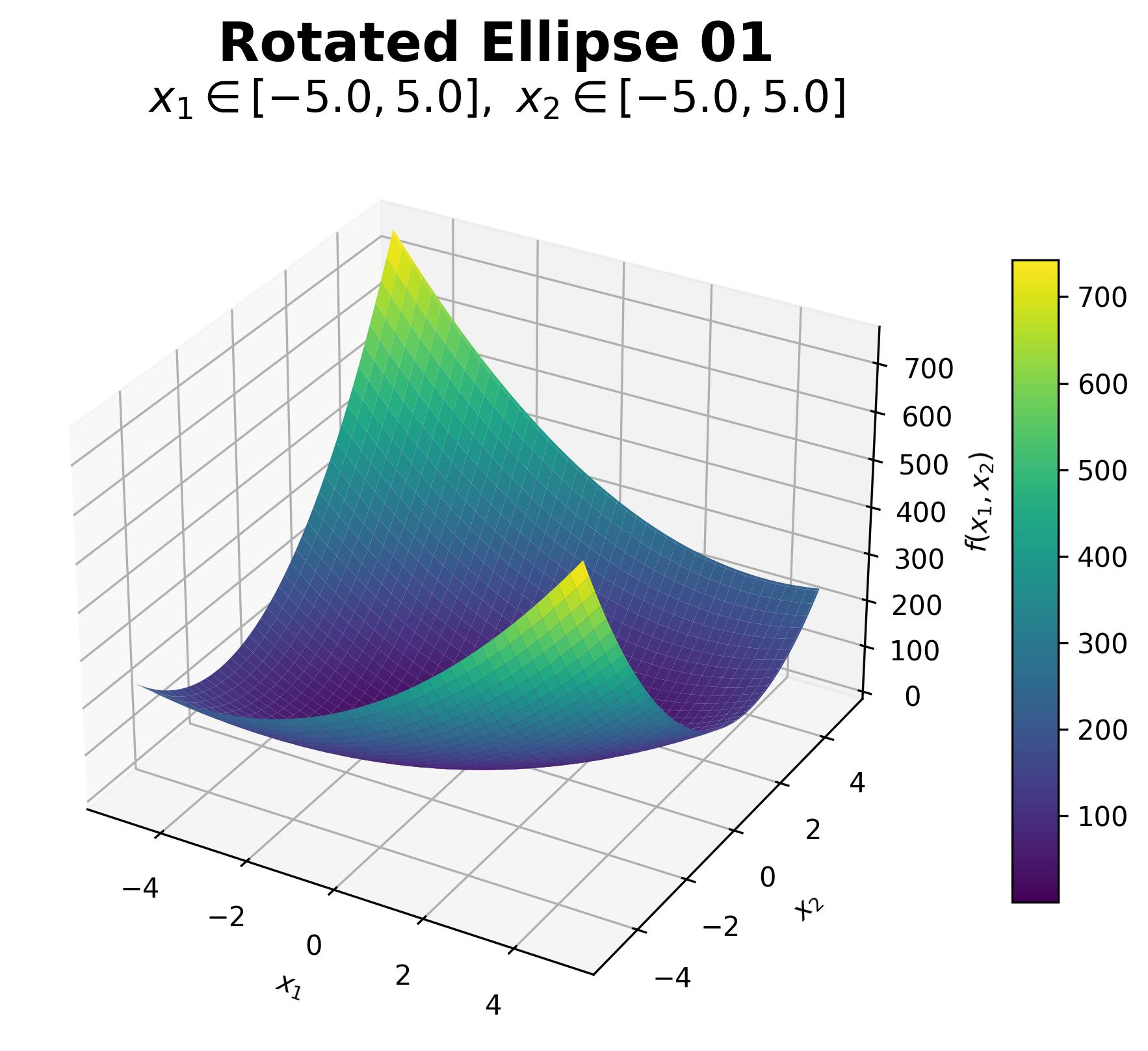} &
\includegraphics[width=0.18\textwidth]{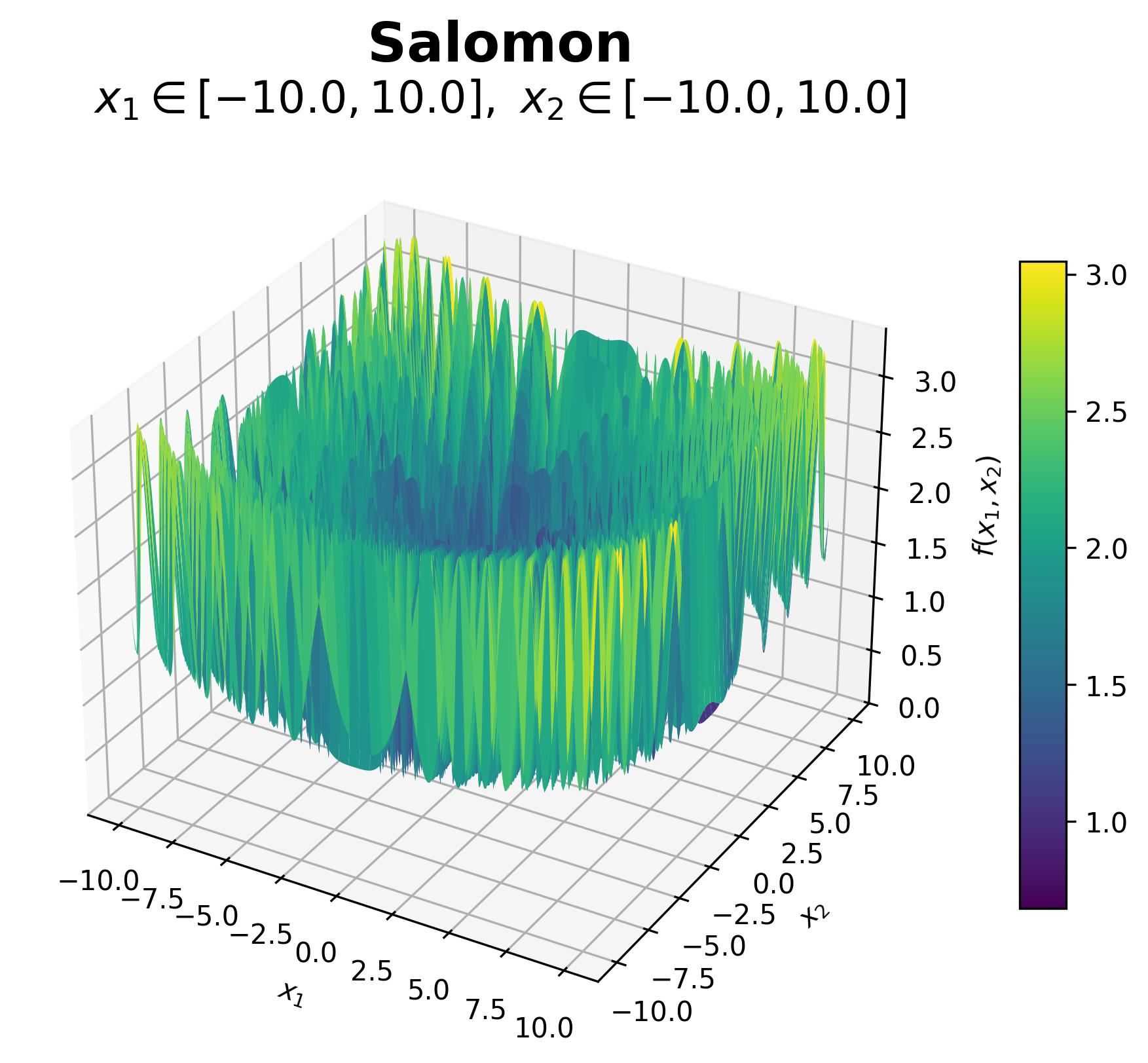} &
\includegraphics[width=0.18\textwidth]{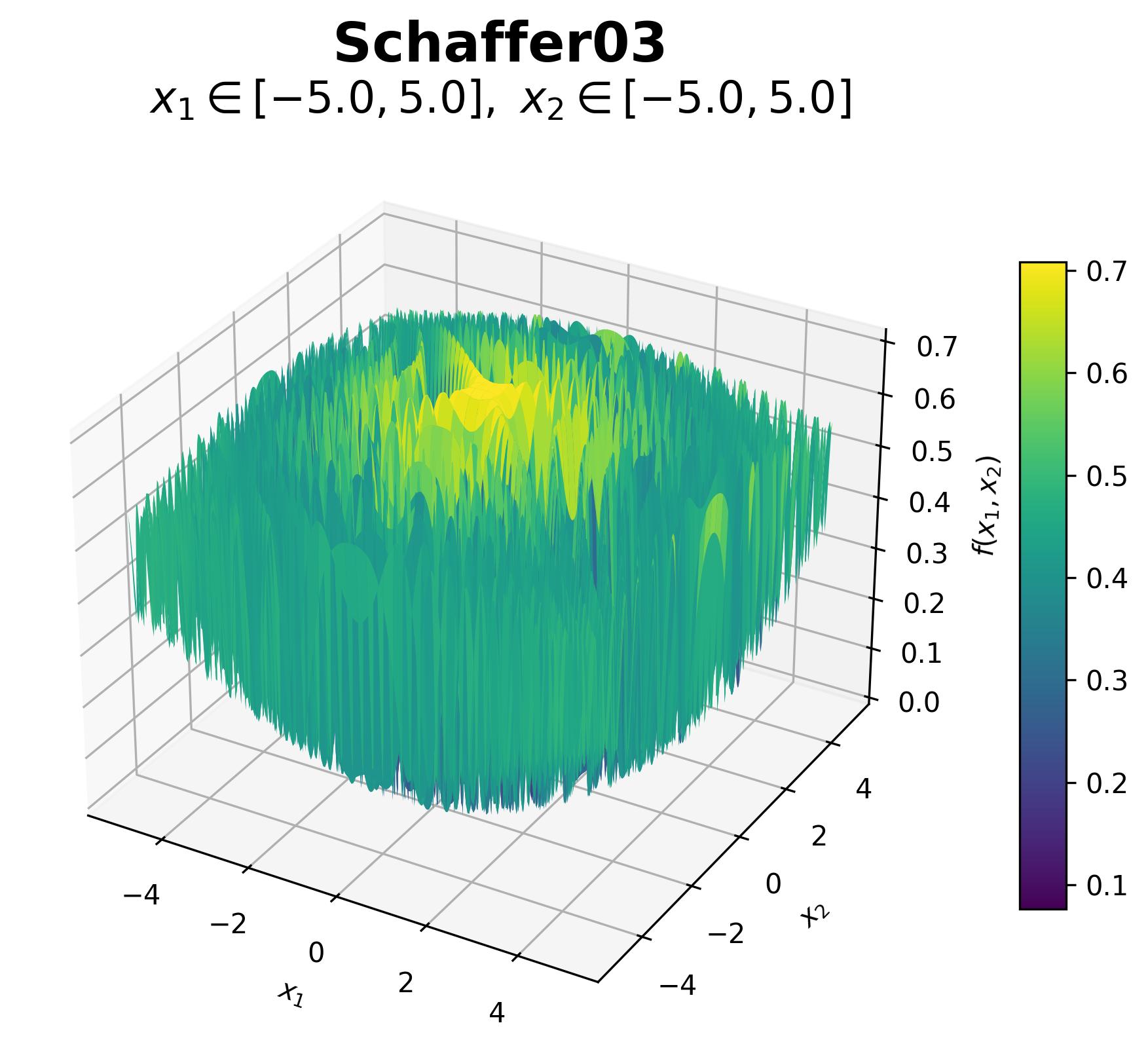} \\
\includegraphics[width=0.18\textwidth]{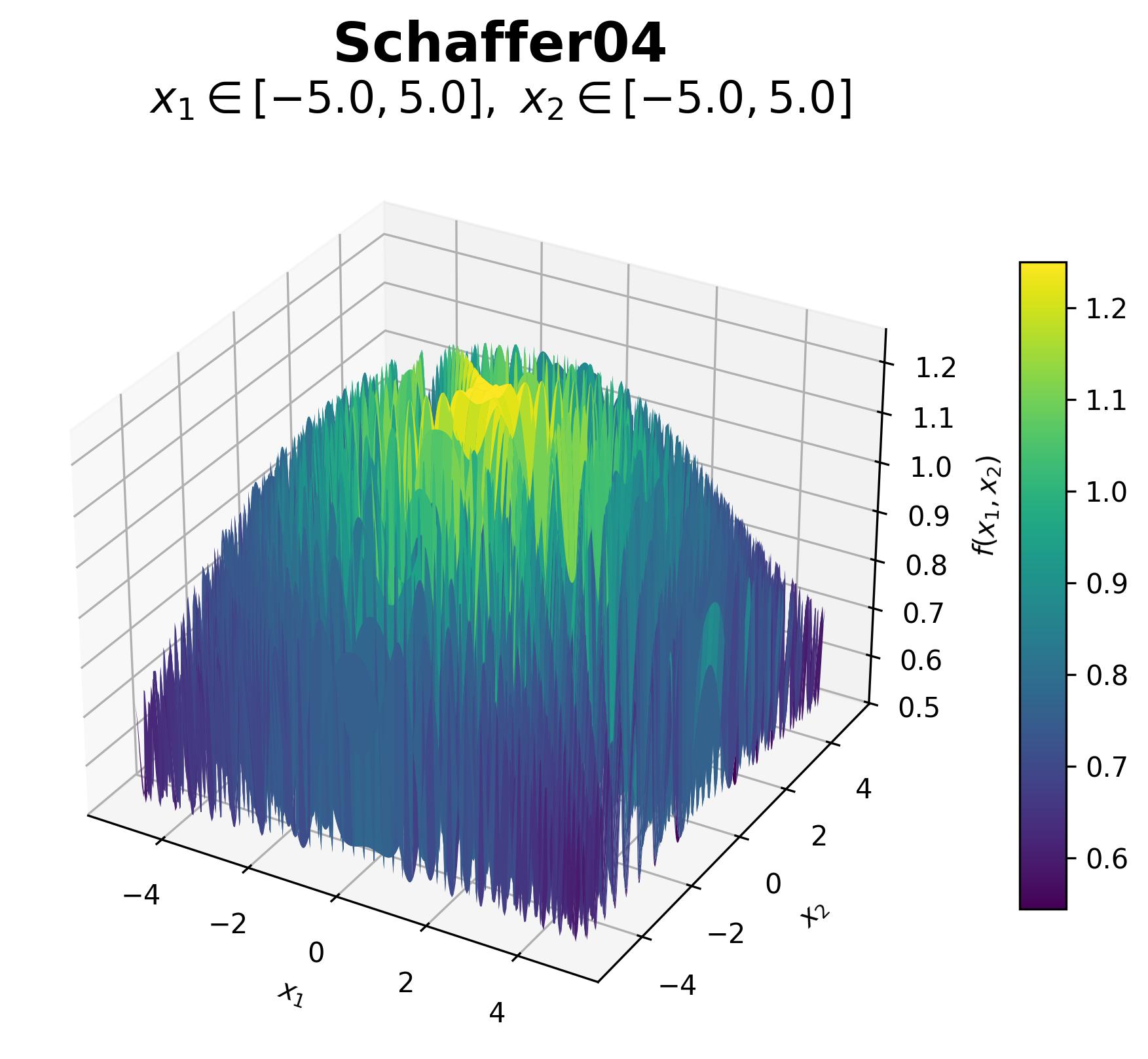} &
\includegraphics[width=0.18\textwidth]{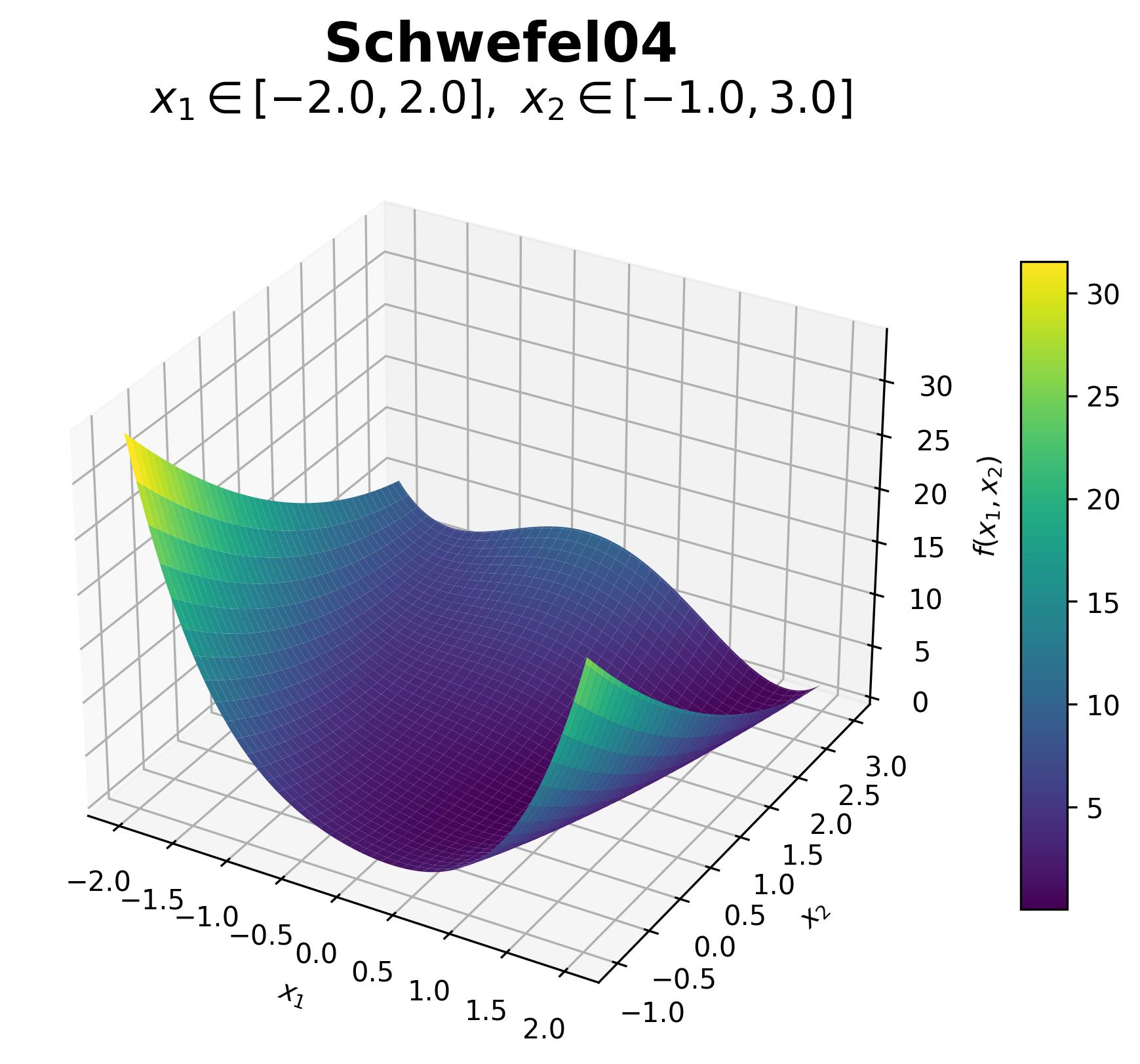} &
\includegraphics[width=0.18\textwidth]{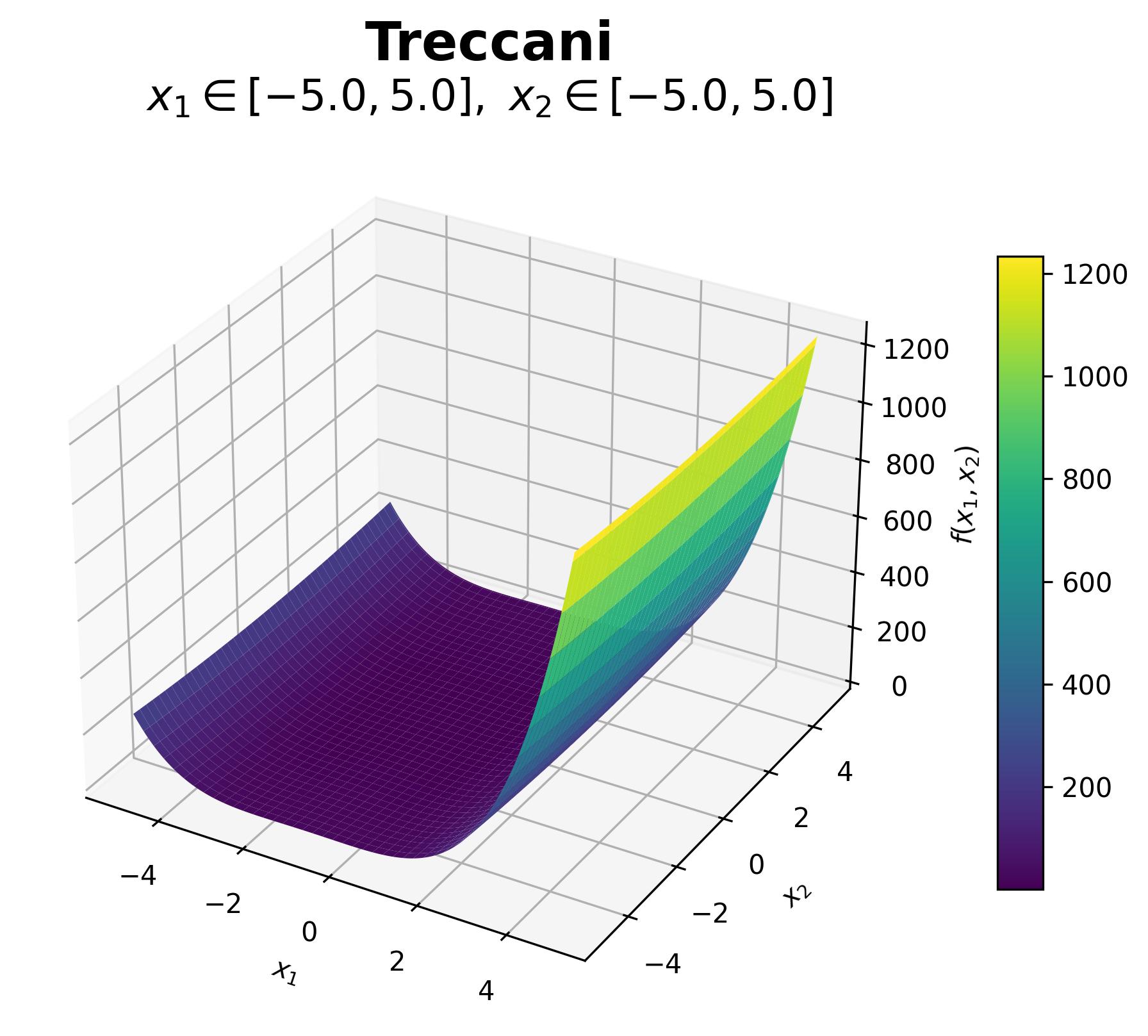} &
\includegraphics[width=0.18\textwidth]{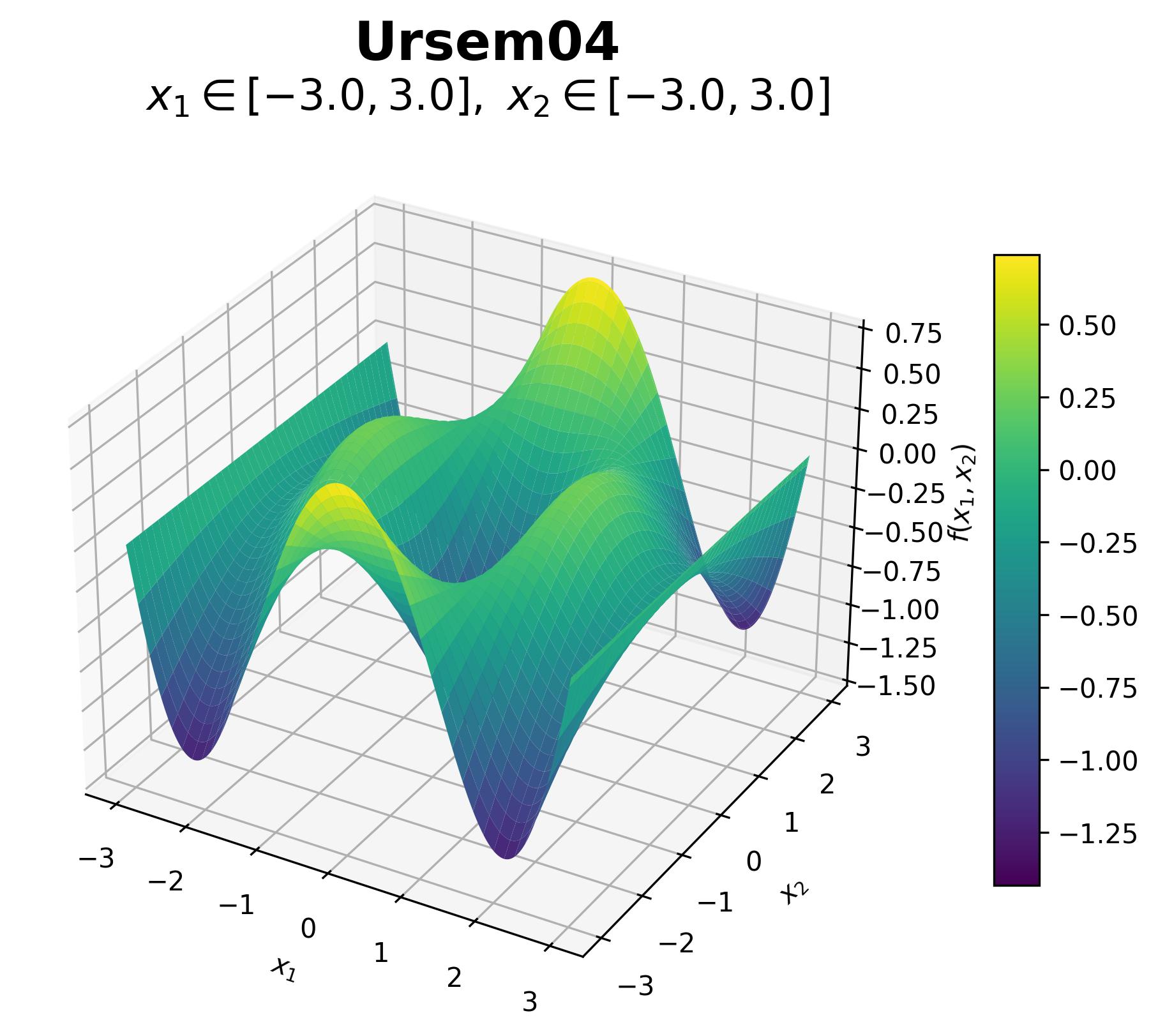} &
\includegraphics[width=0.18\textwidth]{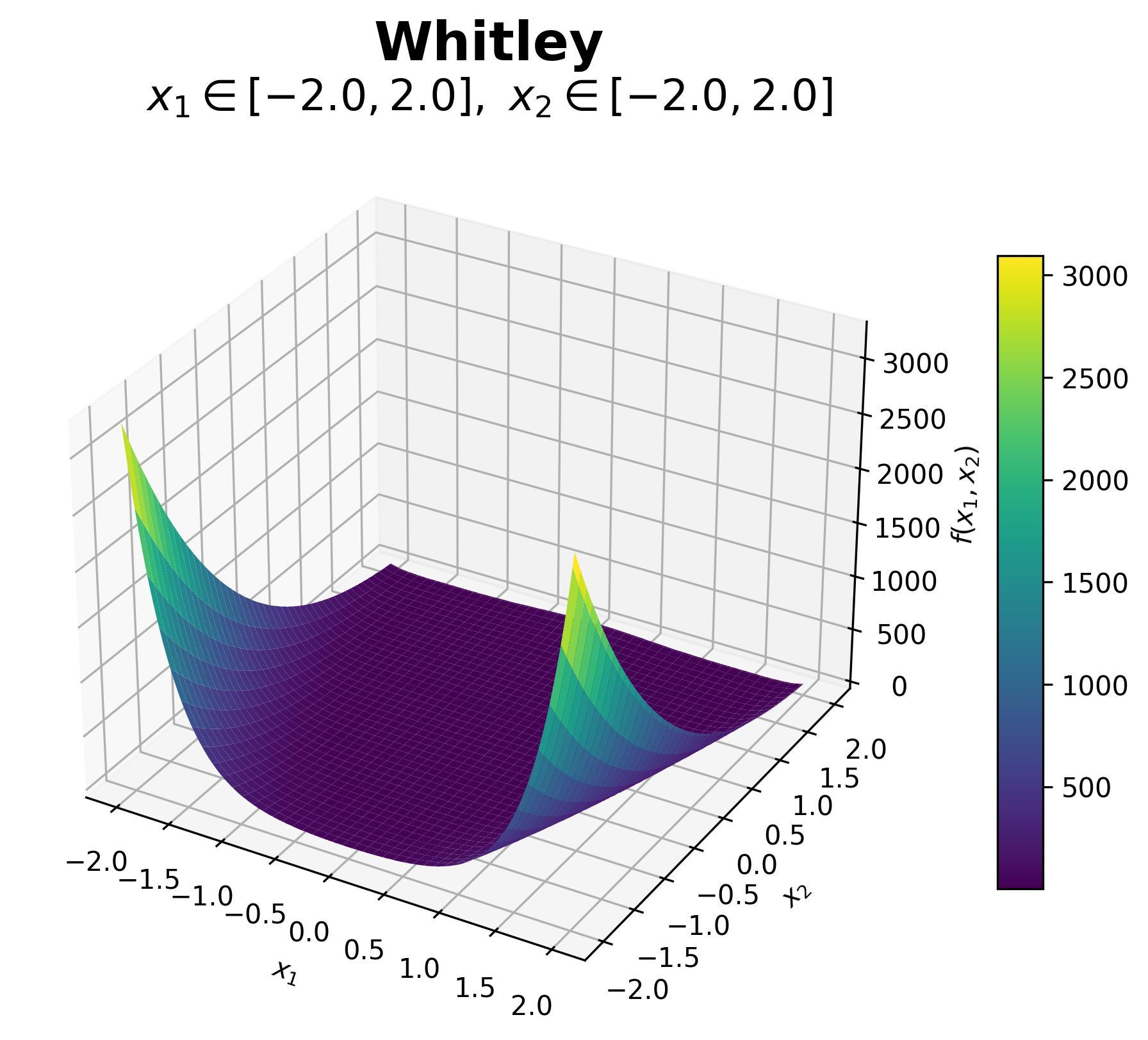} &
\includegraphics[width=0.18\textwidth]{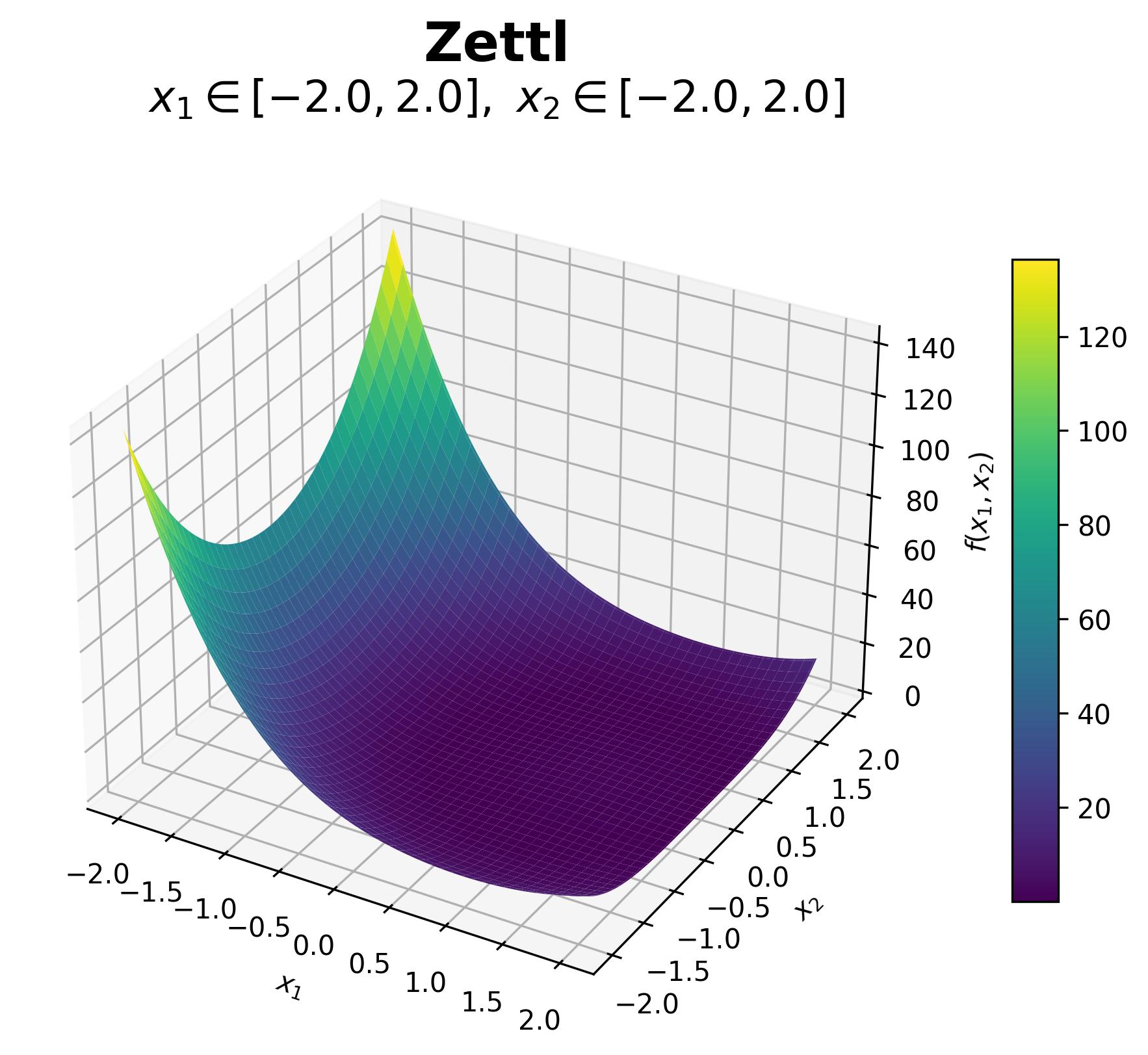} &
\includegraphics[width=0.18\textwidth]{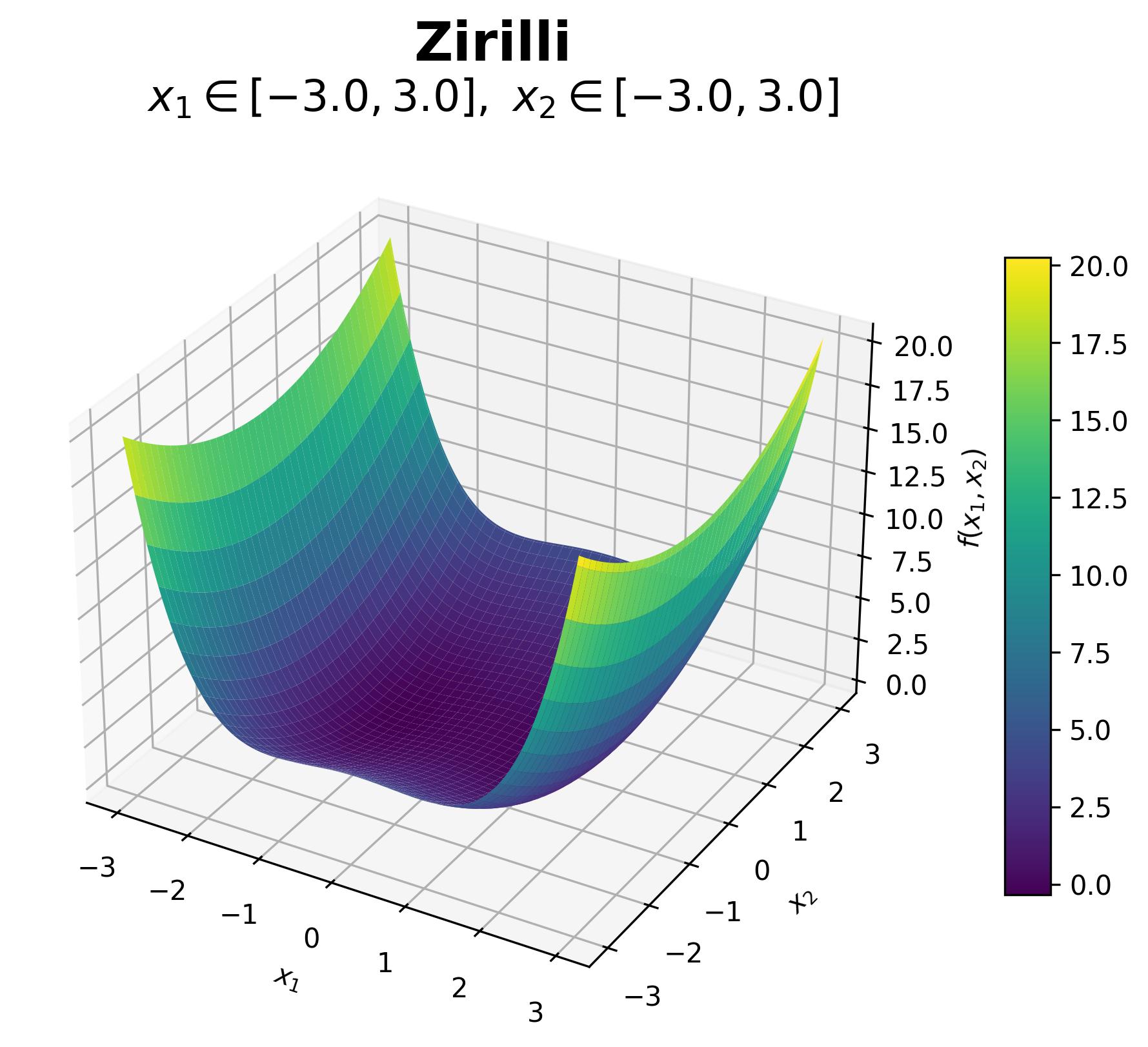} \\
\includegraphics[width=0.18\textwidth]{Zirilli.jpg} \\
\end{tabular}

\caption{Comparative visualisation of 28 benchmark objective functions, highlighting characteristic landscape features—modality, ruggedness, anisotropy, and curvature—relevant to analysing and designing global optimisation methods.}
\label{fig:benchmark_gallery_full}
\end{figure}
\end{landscape}
\printcredits

\bibliographystyle{elsarticle-num-names} 

\bibliography{bibliography}



\end{document}